\documentclass{article}

\usepackage{microtype}
\usepackage{graphicx}
\usepackage{caption}
\usepackage{subcaption}
\usepackage{wrapfig}
\usepackage{booktabs} 

\usepackage{hyperref}

 \usepackage[accepted]{icml2021_oppo}
\usepackage{xspace}

\newcommand{\eg}{\emph{e.g.},\xspace}

\usepackage{amsmath,amsfonts,bm}
\usepackage{amsthm}
\usepackage{thmtools,thm-restate}

\def\Figref#1{Figure~\ref{#1}}

\def\Secref#1{Section~\ref{#1}}

\def\eqref#1{Equation~\ref{#1}}

\def\1{\bm{1}}

\DeclareMathAlphabet{\mathsfit}{\encodingdefault}{\sfdefault}{m}{sl}
\SetMathAlphabet{\mathsfit}{bold}{\encodingdefault}{\sfdefault}{bx}{n}

\def\gN{{\mathcal{N}}}

\newcommand\fakeparagraph[1]{\par\noindent\textbf{{#1}}.\xspace}

\icmltitlerunning{Understanding the effect of sparsity on neural networks robustness}

\begin{document}

\twocolumn[
\icmltitle{Understanding the effect of sparsity on neural networks robustness}



\icmlsetsymbol{equal}{*}

\begin{icmlauthorlist}
\icmlauthor{Lukas Timpl}{equal,tug}
\icmlauthor{Rahim Entezari}{equal,tug,csh}
\icmlauthor{Hanie Sedghi}{goobrain}
\icmlauthor{Behnam Neyshabur}{goo}
\icmlauthor{Olga Saukh}{tug,csh}
\end{icmlauthorlist}

\icmlaffiliation{tug}{Institute of Technical Informatics, Graz University of Technology, Graz, Austria}
\icmlaffiliation{goo}{Blueshift, USA}
\icmlaffiliation{goobrain}{Google Research Brain team, USA}
\icmlaffiliation{csh}{Complexity Science Hub Vienna, Vienna, Austria}

\icmlcorrespondingauthor{Rahim Entezari}{entezari@tugraz.com}
\icmlcorrespondingauthor{Olga Saukh}{saukh@tugraz.at}

\icmlkeywords{sparse networks, robustness, deep learning}

\vskip 0.3in
]



\printAffiliationsAndNotice{\icmlEqualContribution} 

\begin{abstract}
This paper examines the impact of static sparsity on the robustness of a trained network to weight perturbations, data corruption, and adversarial examples. We show that, up to a certain sparsity achieved by increasing network width and depth while keeping the network capacity fixed, sparsified networks consistently match and often outperform their initially dense versions. Robustness and accuracy decline simultaneously for very high sparsity due to loose connectivity between network layers. Our findings show that a rapid robustness drop caused by network compression observed in the literature is due to a reduced network capacity rather than sparsity.
\end{abstract}

\section{Introduction}
\label{sec:intro}

Deep learning methods are increasingly used for solving complex tasks, yet little is known about the choice of the best architecture, the required model size, capacity, and the trade-offs involved. A common strategy is to train overparameterized models and compress them into smaller representations~\cite{hoefler2021sparsity}. This works remarkably well with an almost negligible drop in accuracy~\cite{gale2019state,blalock2020state}, and is crucial to make use of these models in resource-constrained environments. Recent works, however, shows that test accuracy does not capture how model compression impacts the generalization properties of these models~\cite{hooker2020compressed, entezari2019class}.

Related literature refers to robustness as the network generalization ability to small shifts in the distribution that humans are usually robust to. There is a growing body of work studying methods for building robust models. Recent studies~\cite{ShankarRMFRS20,recht2019imagenet} found that image classification models show a consistent accuracy drop when evaluated on ImageNet~\cite{deng2009imagenet} and ImageNetV2~\cite{recht2019imagenet}, while humans achieve the same accuracy. Another line of research aims at minimizing the worst case expected error over a set of probability distributions by applying distributionally robust optimization~\cite{shafieezadehabadeh2015distributionally,duchi2020distributionally,sagawa2020distributionally}. A similar line of work focuses on finding models that have low performance drop on adversarial examples~\cite{Biggio_2018,madry2019deep}.

A recent study by~\citet{hooker2020compressed} shows that model compression, and to a smaller extent quantization, result in tremendous robustness degradation. At the same time, \citet{golubeva2021wider} found that wider networks of the same capacity (same number of parameters) yield better performance. Model compression leads simultaneously to sparser and lower capacity networks, yet the contribution of both effects is mixed. Understanding the impact of these effects on model robustness in isolation is crucial when optimizing machine learning models for resource-constrained devices. This work evaluates the effect of model sparsification while keeping the network capacity, defined by the total number of parameters, fixed.

\begin{figure*}[ht]
    \centering
    \begin{subfigure}[b]{0.325\textwidth}
        \centering
        \includegraphics[width=\linewidth]{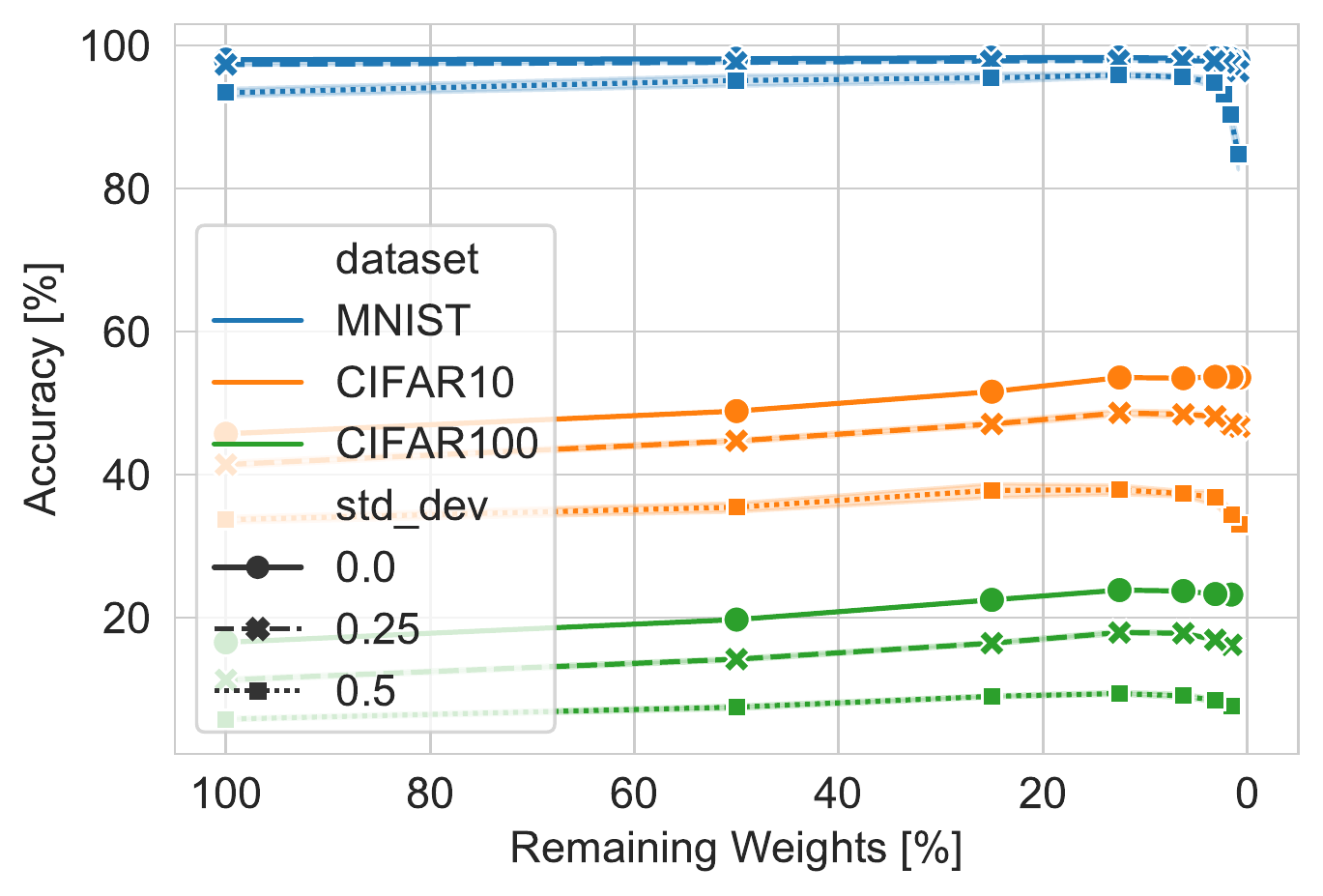}
        \caption{One layer MLP}
        \label{fig:width:perturbated_models_mlp}
    \end{subfigure}
    \hfill
    \begin{subfigure}[b]{0.325\textwidth}
        \centering
        \includegraphics[width=\linewidth]{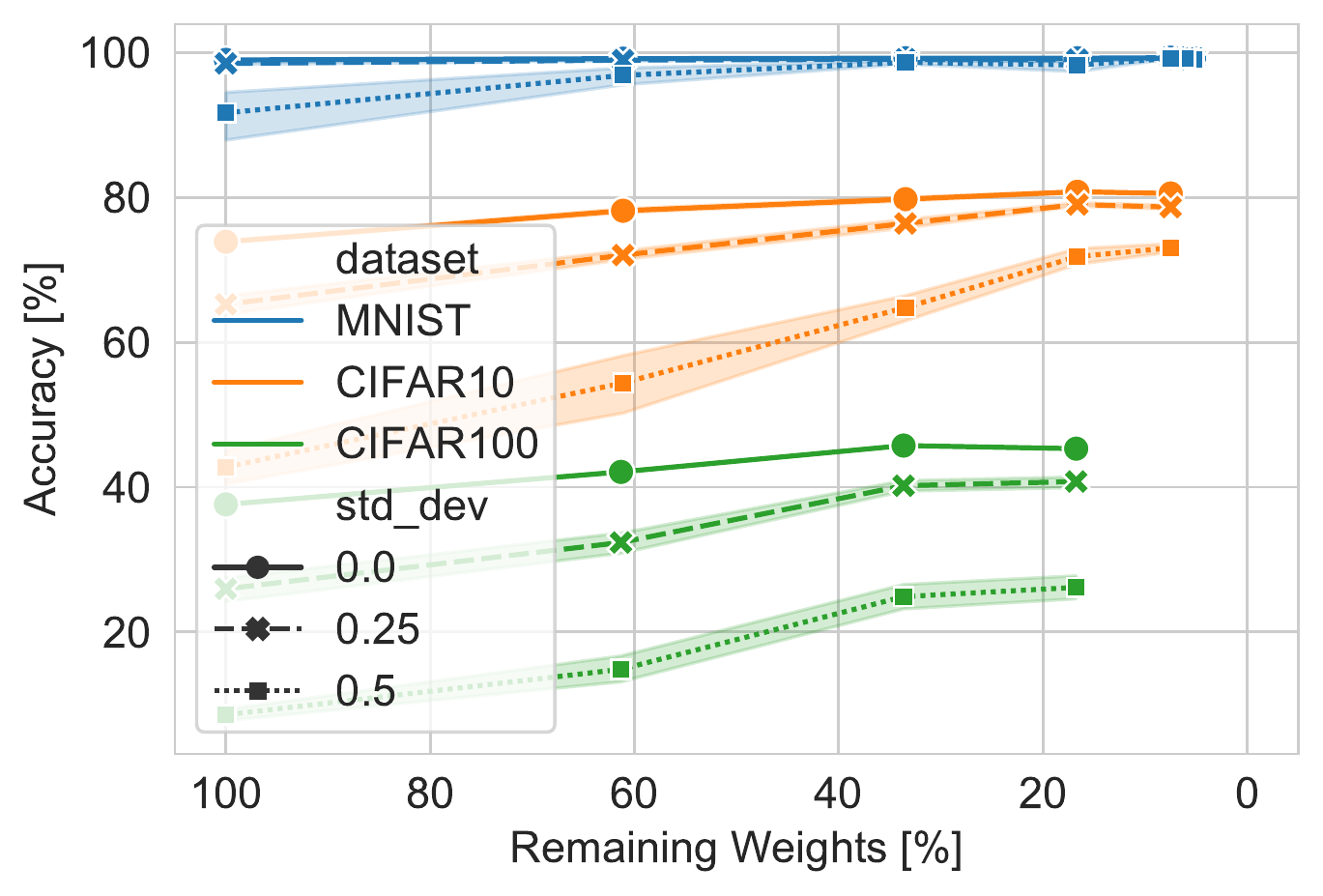}
        \caption{VGG11}
        \label{fig:width:perturbated_models_vgg}
    \end{subfigure}
    \hfill
    \begin{subfigure}[b]{0.325\textwidth}
        \centering
        \includegraphics[width=\linewidth]{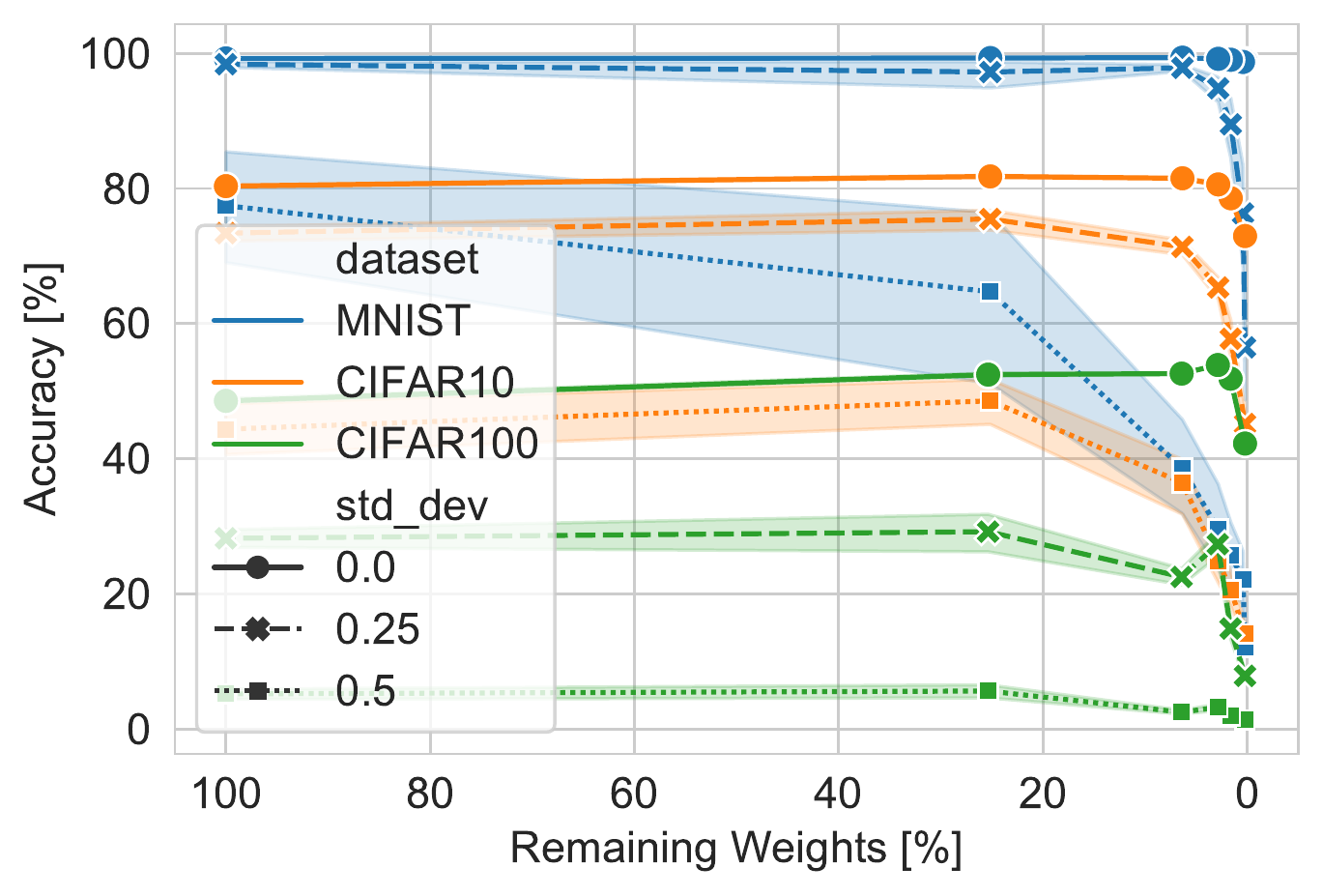}
        \caption{ResNet18}
        \label{fig:width:perturbated_models_resnet}
    \end{subfigure}
    \caption{{\bf Robustness to weight perturbations, sparsification by increasing width.} We add multiplicative Gaussian noise $z_i\sim\gN(\mu, w_i^2\sigma_i^2)$ to each weight and evaluate model performance. We observe that as we move towards higher sparsity levels, the performance first increases then decreases in extreme sparsity levels. We note that such increase is happening earlier for simpler tasks like MNIST. This performance improvement indicates a flatter loss landscape around the minima suggesting better generalization.}
    \label{fig:width:perturbated_models}
    \end{figure*}

\begin{figure*}[ht]
    \centering
    \begin{subfigure}[b]{0.33\textwidth}
        \centering
        \includegraphics[width=\linewidth]{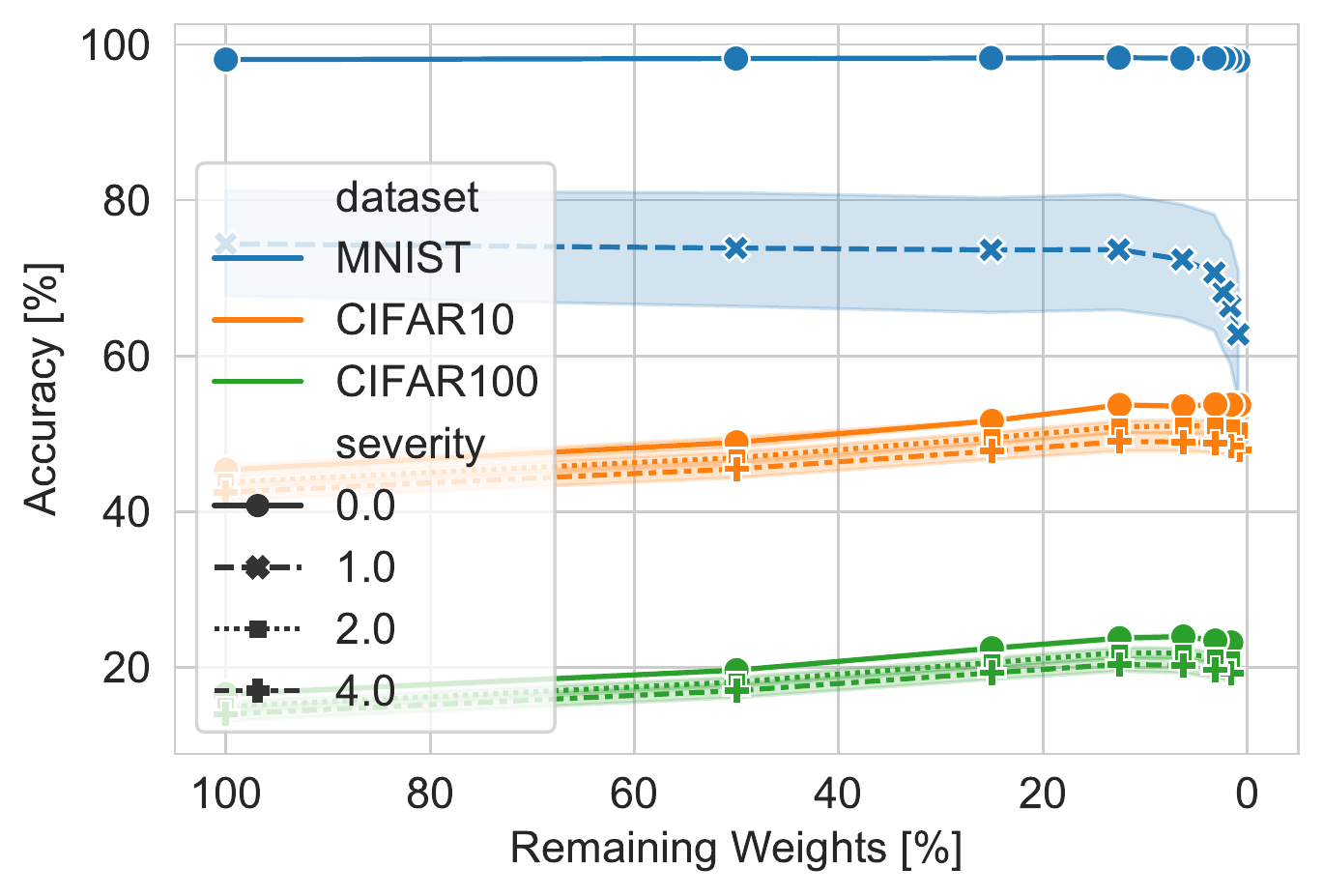}
        \caption{One layer MLP}
        \label{fig:width:corrupted_data_mlp}
    \end{subfigure}
    \hfill
    \begin{subfigure}[b]{0.33\textwidth}
        \centering
        \includegraphics[width=\linewidth]{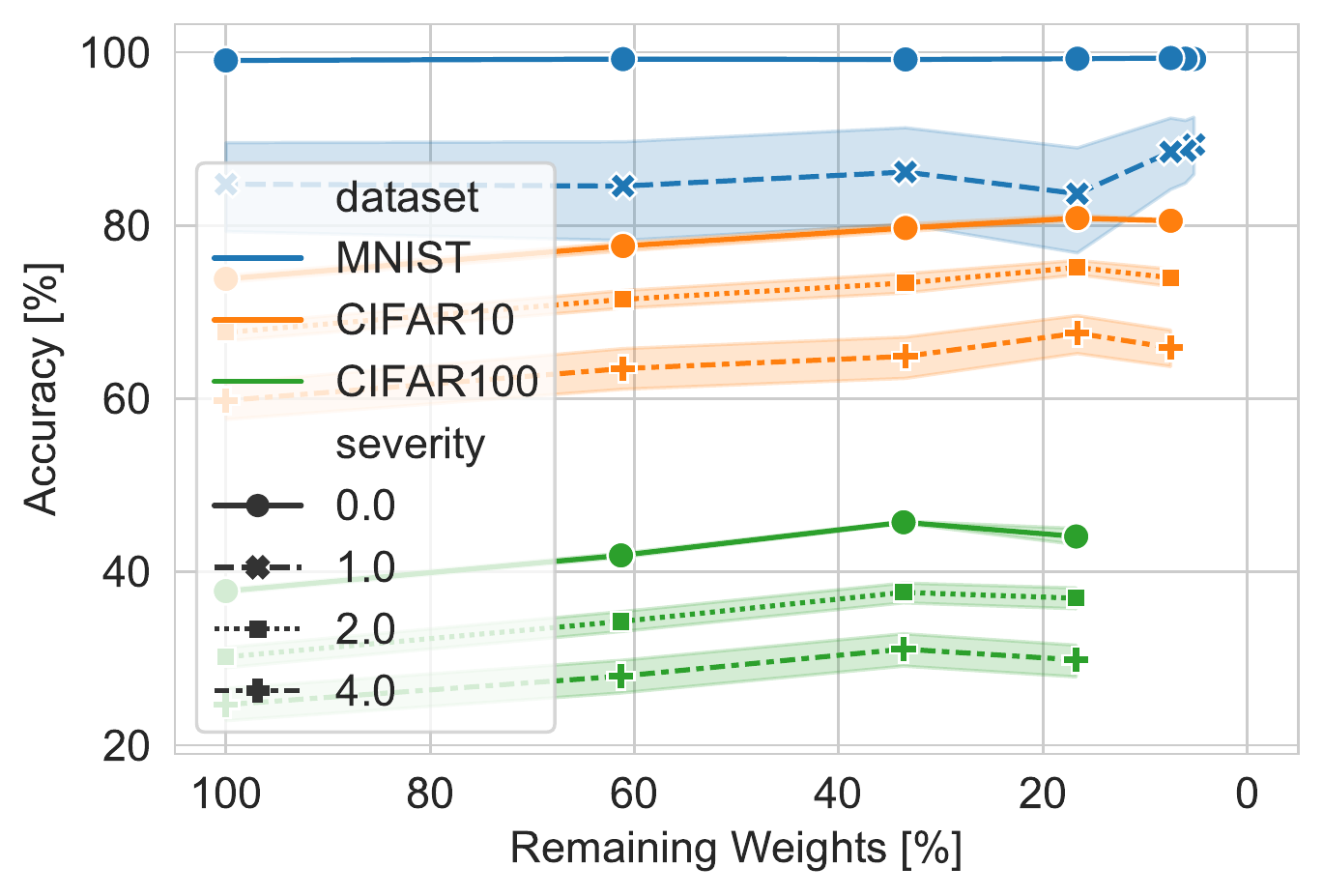}
        \caption{VGG11}
        \label{fig:width:corrupted_data_vgg}
    \end{subfigure}
    \hfill
    \begin{subfigure}[b]{0.33\textwidth}
        \centering
        \includegraphics[width=\linewidth]{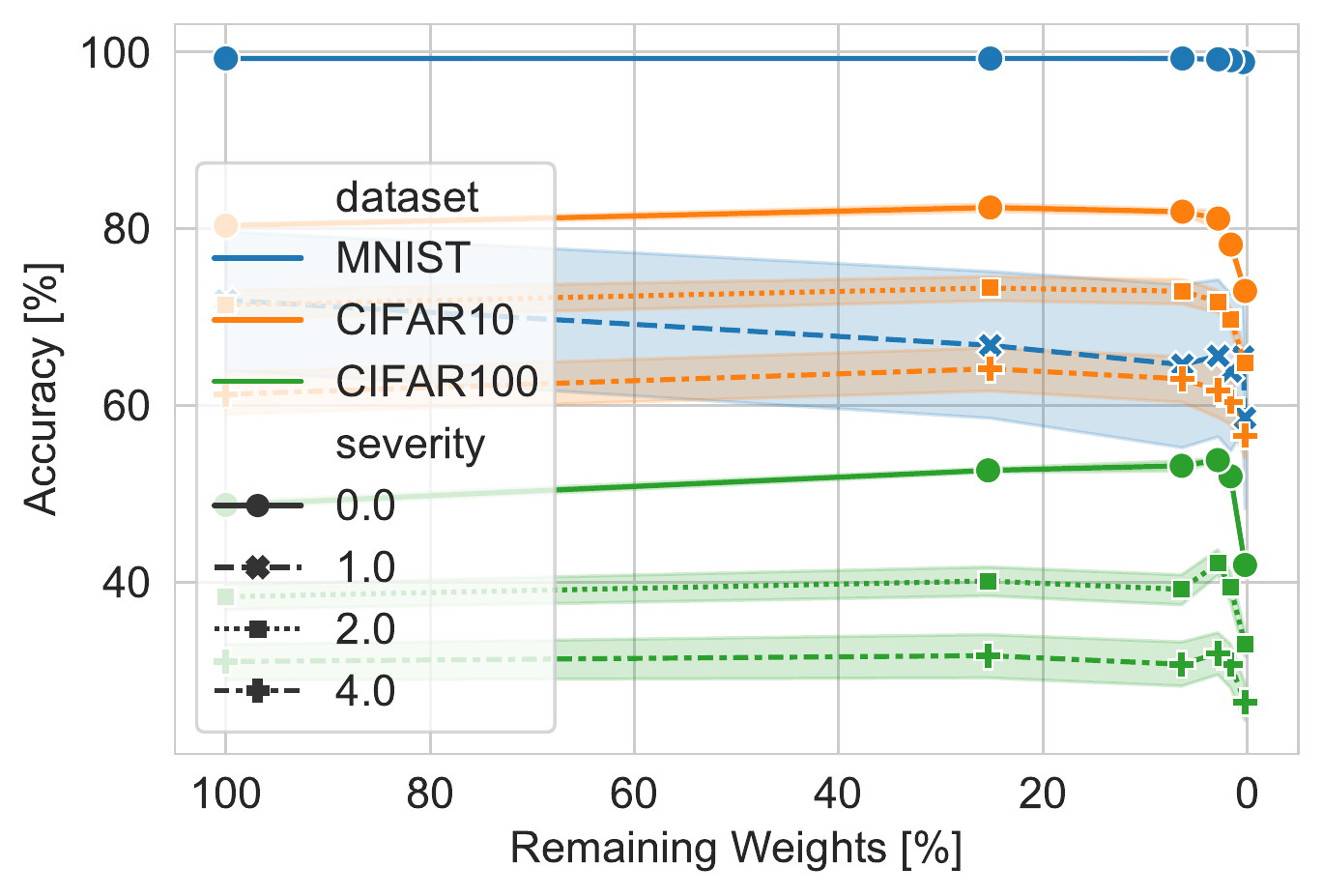}
        \caption{ResNet18}
        \label{fig:width:corrupted_data_resnet}
    \end{subfigure}
    \caption{{\bf Robustness to data corruption, sparsification by increasing width.} We evaluate the performance of the models on corrupted datasets MNIST-C, CIFAR10-C and CIFAR100-C. We observe that as we move towards higher sparsity levels, the performance first increases then decreases in extreme sparsity levels. We note that such increase is happening earlier for simpler tasks like MNIST.}
    \label{fig:width:corrupted_data}
\end{figure*}

\begin{figure*}[htb]    
    \begin{subfigure}[b]{0.33\textwidth}
        \centering
        \includegraphics[width=\linewidth]{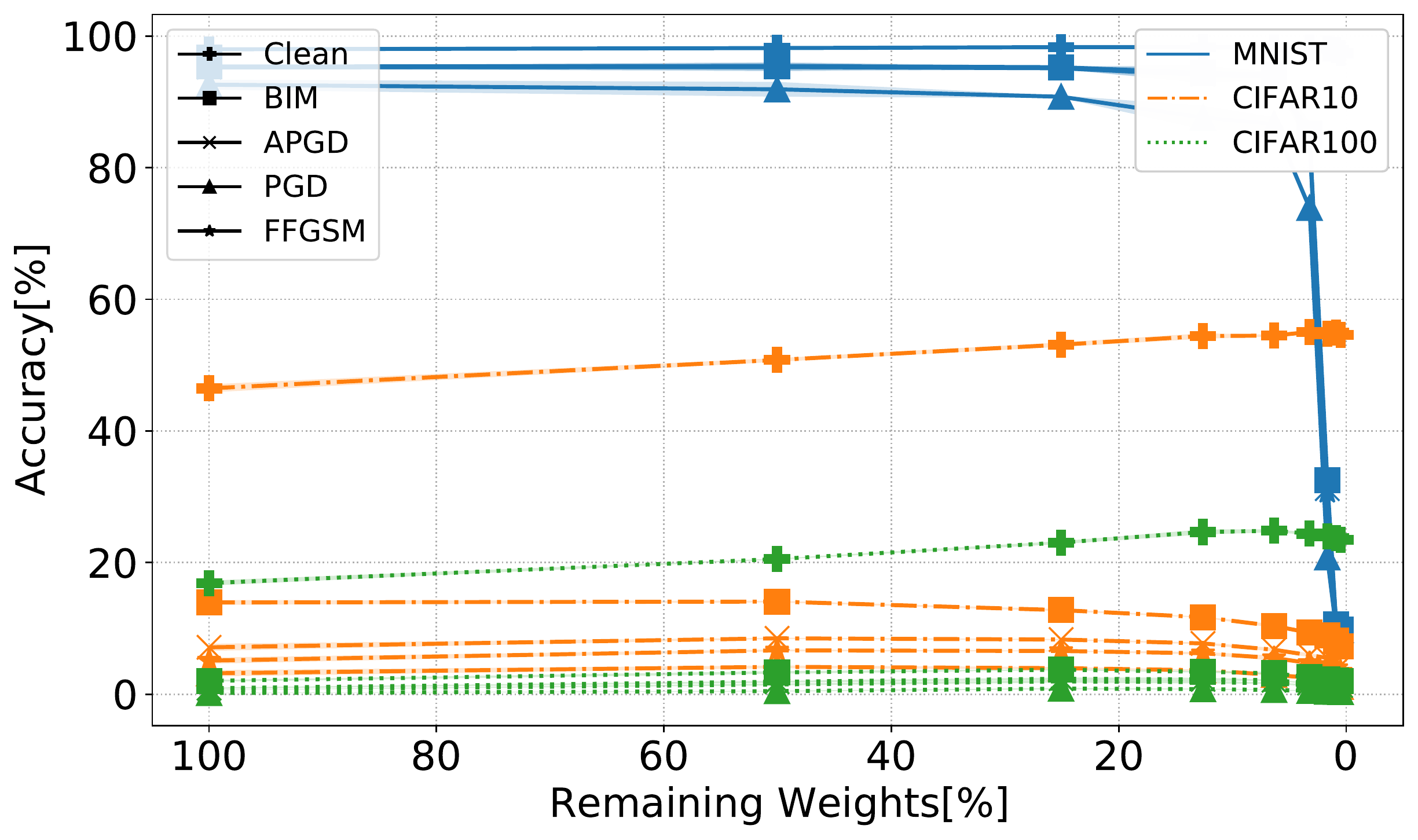}
        \caption{One layer MLP}\label{adv_mlp_1layer}
    \end{subfigure}
    \hfill
    \begin{subfigure}[b]{0.33\textwidth}
        \centering
        \includegraphics[width=\linewidth]{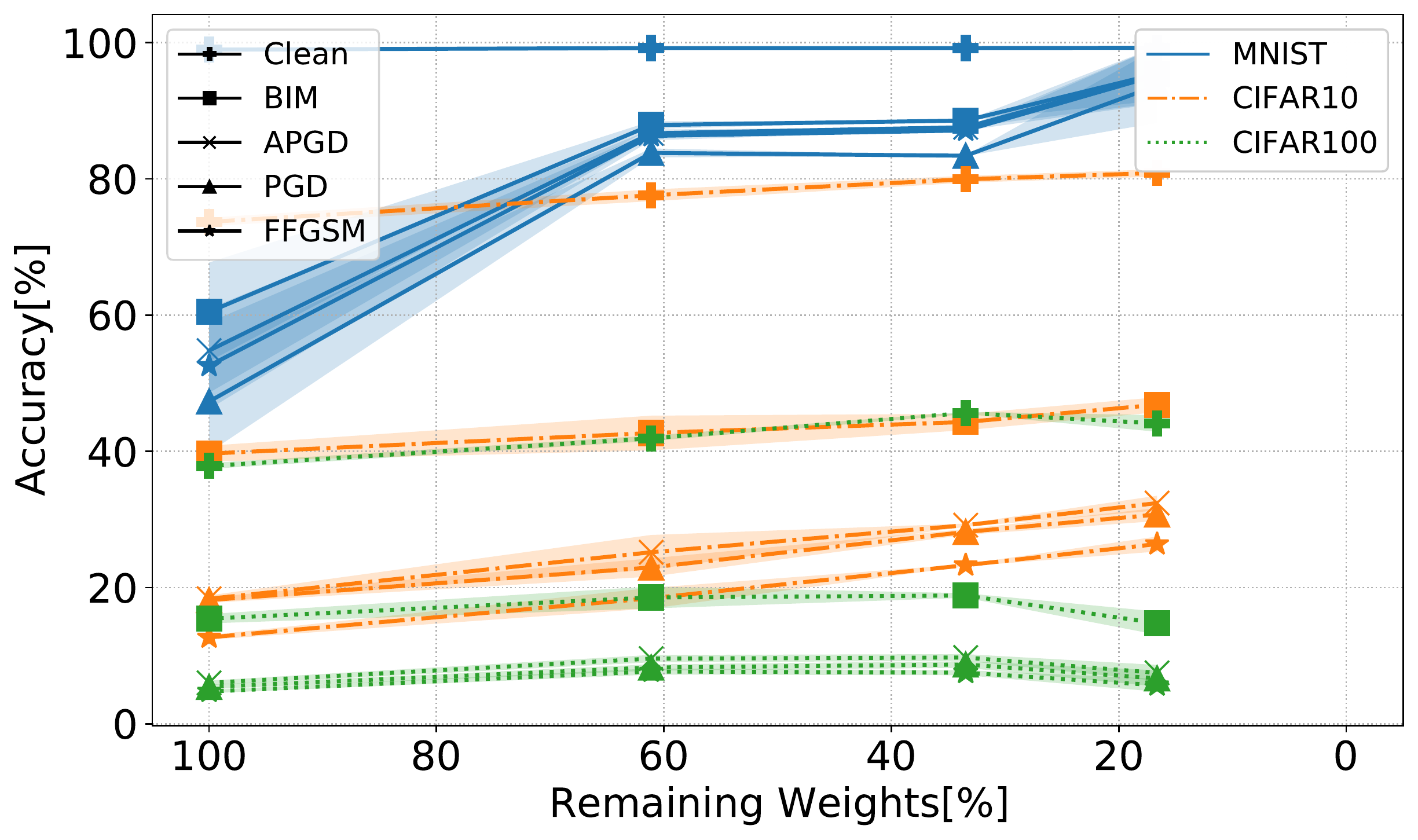}
        \caption{VGG11}\label{adv_vgg11}
    \end{subfigure}
    \hfill
    \begin{subfigure}[b]{0.33\textwidth}
        \centering
        \includegraphics[width=\linewidth]{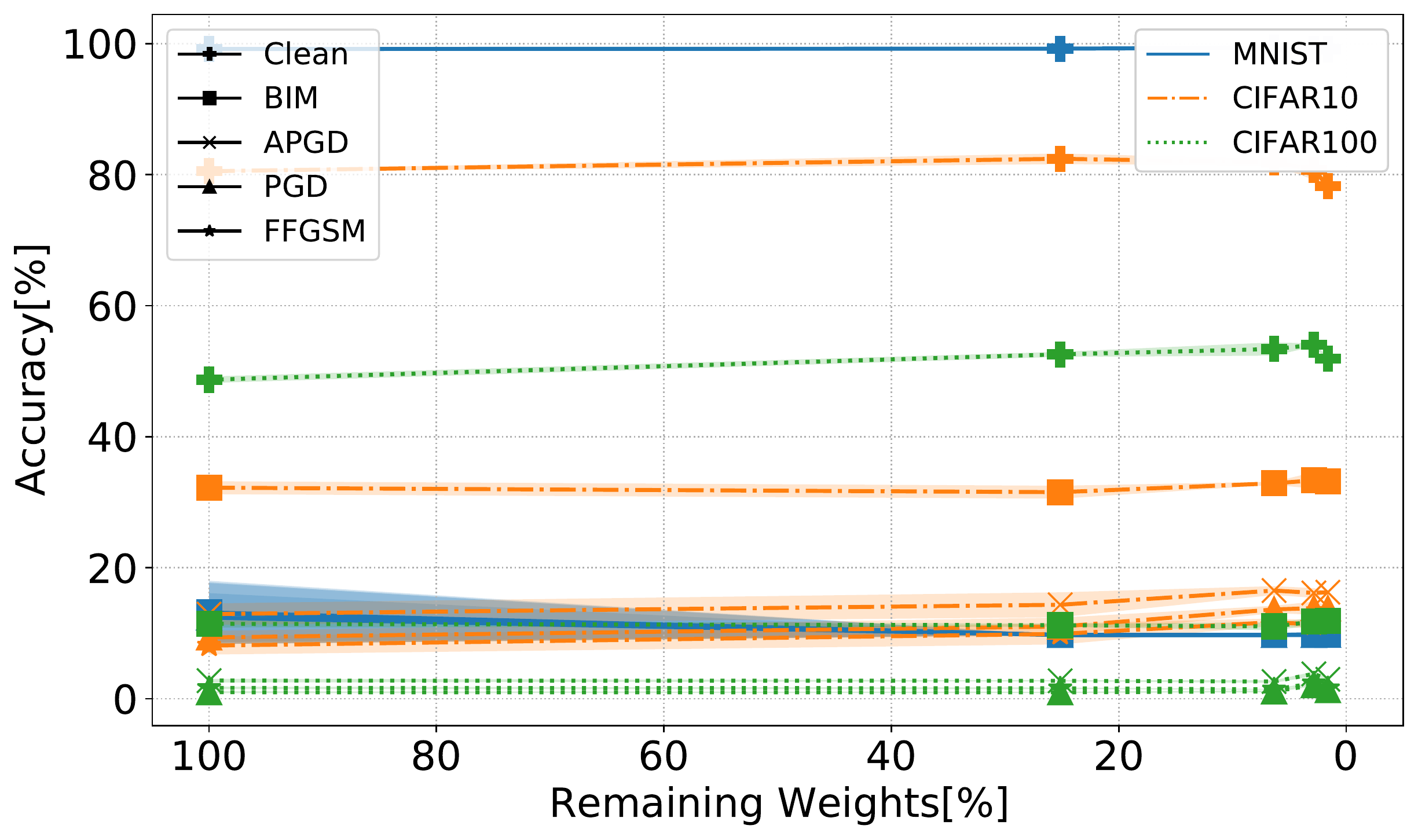}
        \caption{ResNet18}
        \label{adv_resnet18}
    \end{subfigure}
    \caption{\small {\bf Robustness to adversarial attacks. Sparsification by increasing width.} Robustness to all adversarial attacks (BIM~\cite{kurakin2016adversarial}, APGD~\cite{croce2020reliable}, PGD~\cite{madry2019deep}, FFGSM~\cite{goodfellow2014explaining}) is improved as we have less remaining weights and decreases for extreme sparsity levels where overall network accuracy (clean) drops.}
    \label{fig:adv_corrupted_data_selected}
\end{figure*}

\begin{figure}[!ht]
    \centering
    \begin{subfigure}[b]{0.49\linewidth}
        \centering
        \includegraphics[width=\linewidth]{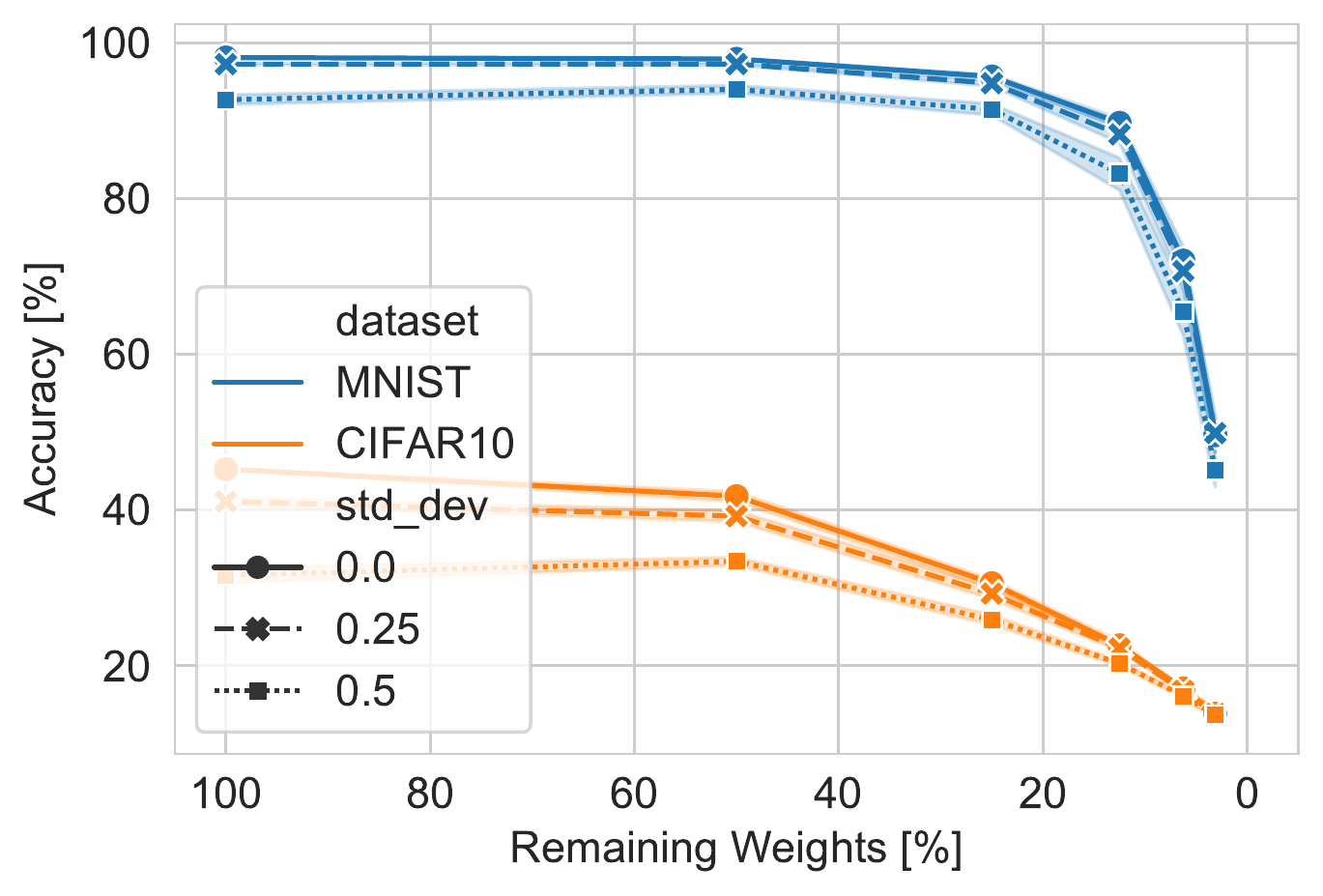}
        \caption{One layer MLP}
        \label{fig:posttraining:perturbated_models_mlp}
    \end{subfigure}
    \hfill
    \begin{subfigure}[b]{0.49\linewidth}
        \centering
        \includegraphics[width=\linewidth]{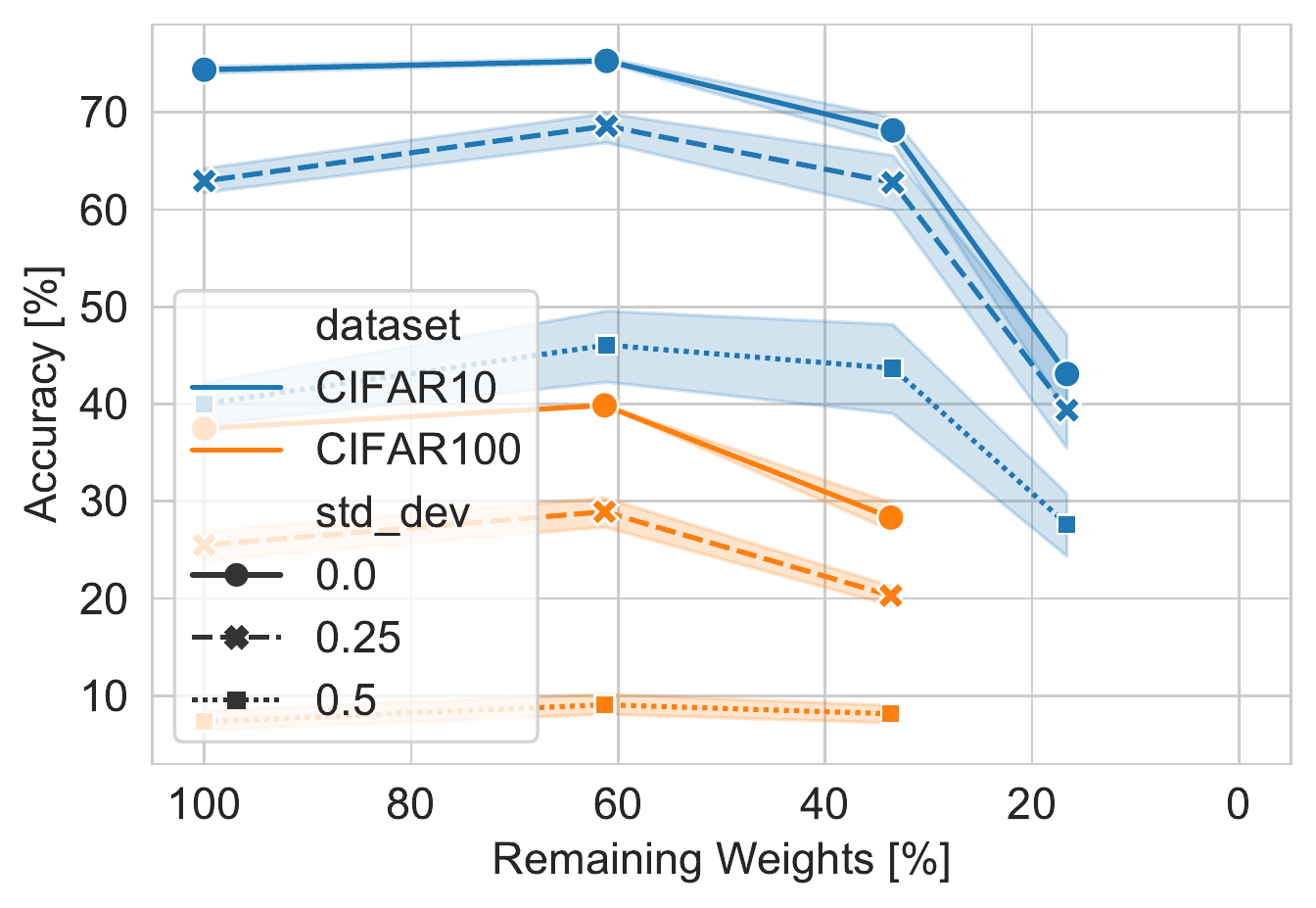}
        \caption{VGG11}
        \label{fig:posttraining:perturbated_models_vgg}
    \end{subfigure}
    \caption{{\bf Robustness to weight perturbations. Sparsification after training by increasing width.} We add multiplicative Gaussian noise $z_i\sim\gN(\mu, w_i^2\sigma_i^2)$ to each weight and evaluate performance on test data. As we move towards higher sparsity levels, the performance decreases in extreme sparsity levels.}
    \label{fig:posttraining:perturbated_models}
\end{figure}

\begin{figure}[!ht]
    \centering
    \begin{subfigure}[b]{0.49\linewidth}
        \centering
        \includegraphics[width=\linewidth]{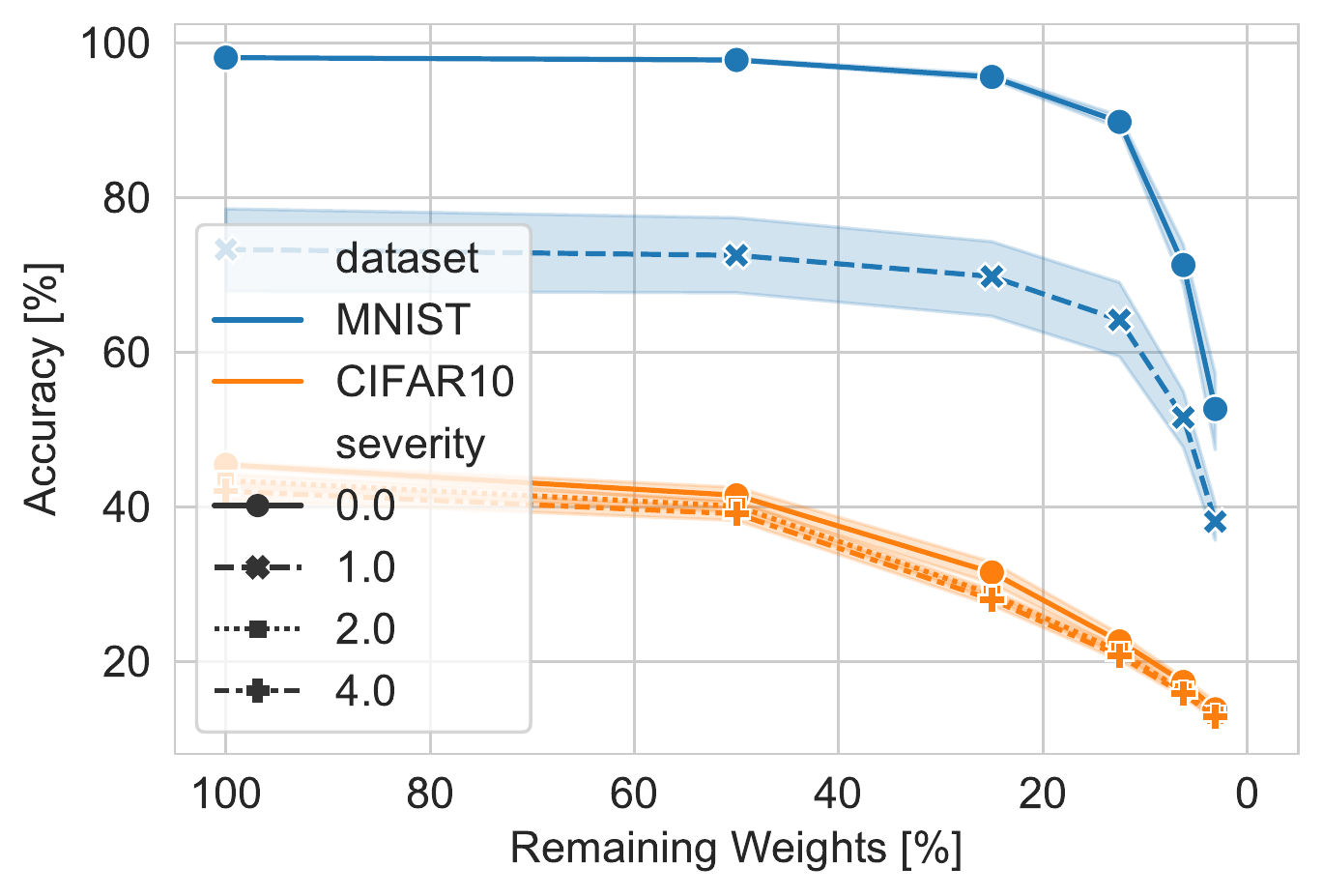}
        \caption{One layer MLP}
        \label{fig:posttraining:corrupted_data_mlp}
    \end{subfigure}
    \hfill
    \begin{subfigure}[b]{0.49\linewidth}
        \centering
        \includegraphics[width=\linewidth]{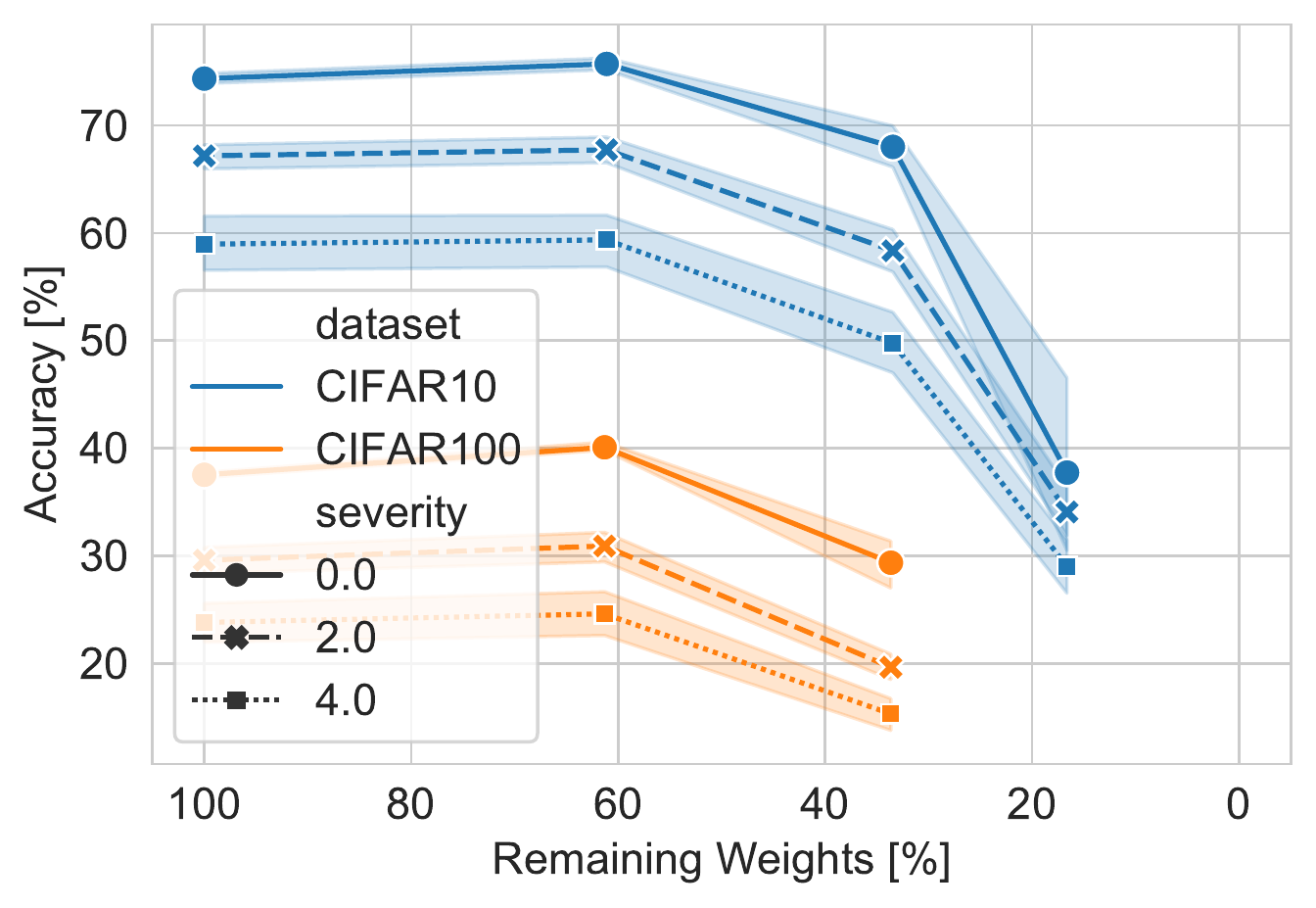}
        \caption{VGG11}
        \label{fig:posttraining:corrupted_data_vgg}
    \end{subfigure}
    \caption{{\bf Robustness to data corruption. Sparsification after training by increasing width.} We evaluate on the corrupted datasets MNIST-C, CIFAR10-C and CIFAR100-C. sompare to static sparsity at the prior to training, robustness degrades sooner.}
    \label{fig:posttraining:corrupted_data}
\end{figure}

\fakeparagraph{Contributions}
We hypothesise that sparsity alone does not hurt model robustness when the network capacity is fixed and provide empirical evidence to support this hypothesis in a number of settings. We run our study on a range of network architectures (MLPs, VGG and ResNets), datasets (MNIST, CIFAR-10, CIFAR-100), robustness tests (weight perturbations, data corruptions, adversarial examples) and evaluate the overall and per class network performance. We observe that for randomly initialized models with a static sparsity pattern applied before or after training, network sparsification does not hurt or even improves robustness to a certain sparsity compared to a dense network of the same capacity. Robustness and accuracy decline simultaneously for very high sparsity due to loose connectivity between network layers. We show that our hypothesis holds when introducing sparsity by increasing network width and depth in separate experiments, applied before and after training. These findings show that a rapid robustness drop caused by network compression observed in the literature is due to a reduced network capacity rather than sparsity.

\section{Experimental Framework}
\label{sec:framework}
We hypothesise that sparsity, while keeping the number of parameters fixed, does not hurt network robustness. We support our hypothesis by exhaustive tests covering multiple datasets, network architectures, model and data corruptions, sparsity levels, sparsification methods and schedules. The details are given below.

\fakeparagraph{Datasets and architectures} 
The datasets used in the experiments include MINST~\cite{mnist}, CIFAR-10~\cite{cifar100}, and CIFAR-100)~\cite{cifar100}. We fix the number of weights in each network architecture (one layer MLP, VGG16~\cite{simonyan2014very}, ResNet18~\cite{he2015deep}) throughout all experiments, by increasing the width or depth and introducing the proper corresponding sparsity. See \textit {sparsification methods} for more details. We use one layer MLP with $2^7$ hidden units, VGG with 11 layers, 
and ResNet18 as base architectures.
We refer to these vanilla architectures as to 100\,\%-networks {\it before} sparsification. Note that for both ResNet and VGG our vanilla implementation uses the layer width of 16 as the base architecture, which is lower than 64 used in the original architecture. We use width to set the number of output channels for the first layer and use the same width ratios as the respective vanilla architectures for the following layers. All networks were trained using SGD with momentum 0.9. Details for each model family are provided in \Secref{sec:implementationdetails}.

\fakeparagraph{Sparsification methods}\label{par:sparsification}
Existing literature covers multiple ways to make use of sparsity during and after model training including static and dynamic sparsity (\eg $\beta$-Lasso~\cite{neyshabur2020towards}), iterative hard thresholding (\eg Lottery Ticket Hypothesis with various pruning strategies~\cite{frankle2018lottery, renda2020comparing}) and others. \cite{hoefler2021sparsity} provides a comprehensive survey on pruning strategies. Sparsification without changing the number of parameters was investigated in~\cite{golubeva2021wider}. In their study static sparsity showed the most prominent impact on network performance and is thus adopted in this work.

We sparsify a network while preserving its capacity by changing the network's width or depth. When sparsifying by increasing width, we leverage the approach introduced in  \cite{golubeva2021wider}: every layer of the network is sparsified by removing weights at random in proportion to the layer size, using a static mask generated at initialization. This approach is referred to as \emph{static sparsity}. We build on its publicly available  implementation~\cite{golubeva2021wider}. Sparsifying by increasing network depth involves duplicating layers and then applying a static random mask to sparsify the weight tensors. When sparsifying by increasing depth, we consider MLP with $2^9$ hidden units in each layer, and add layers of the same size. For VGG and ResNet we build architecture families VGG11, VGG13, VGG16 and ResNet18, ResNet34, ResNet50 all enjoying the default width of 64.

\fakeparagraph{Sparsification schedules}
In addition to static sparsity applied prior to network training, we also investigate network pruning after training by removing a certain amount of weights with the lowest magnitude to match the required sparsity level. Note that no fine-tuning is applied.

\fakeparagraph{Robustness measures}
We evaluate the impact of sparsity on model performance with respect to weight perturbations~\cite{vonoswald2021neural},  data corruptions~\cite{hendrycks2019robustness} and natural adversarial examples~\cite{hendrycks2021natural}.

{\it Model perturbation.} Similarly to \cite{vonoswald2021neural}, we perturb model weights by applying Gaussian noise $z_i\sim\gN(\mu,w_i^2\sigma_i^2)$ in proportion to the magnitude of each weight $w_i, i \in L$, 
and then measure the difference in the loss $\delta \mathcal{L} = {\mathbb{E}_z}[\mathcal{L}(w_i+z)-\mathcal{L}(w_i)]$.
Accuracy drop due to model perturbation is related to the flatness of the loss landscape around the obtained optimum.
Robustness to weight perturbation could also represents a proxy for quantization error. This error is introduced in neural network compression by weight quantization in the literature~\cite{novac2021quantization}. 

{\it Corrupted data.} We apply numerous algorithmically generated corruptions, similar to the ones evaluated in \cite{hooker2020compressed} (\eg blur, contrast, pixelation) to all datasets used in this paper. This allows us investigating how sensitive the sparsified models are to data corruptions of different severity which humans are oblivious to. Our corrupted datasets are MNIST-C \cite{mu2019mnist}, CIFAR10-C and CIFAR100-C \cite{hendrycks2019robustness}.

{\it Natural adversarial examples.} We use Torchattacks~\cite{kim2020torchattacks} to generate a diverse range of adversarial attacks for different combination of mentioned architectures and datasets. This include FGSM~\cite{goodfellow2014explaining}, BIM~\cite{kurakin2016adversarial}, APGD~\cite{croce2020reliable}, and PGD~\cite{madry2019deep}.

When applying sparsity, we evaluate both the overall model performance and its performance on the most sensitive class. We follow the methodology introduced in \cite{hooker2020compressed} and evaluate the change to class level recall compared to the overall model accuracy. The obtained results are presented below.

\section{Results}
\label{sec:results}

\fakeparagraph{Perturbed model weights}
We first investigate the networks that were sparsified while growing the width to keep their capacity fixed. \Figref{fig:width:perturbated_models} shows that as we move towards higher sparsity levels, the test performance first increases then decreases in extreme sparsity levels. We note that such increase is happening earlier for simpler tasks like MNIST. We observe that sparse configurations
are indeed in flatter regions of weight space as $\delta\mathcal{L}$ increases more slowly with $\delta{z_i}$. This suggests better robustness and generalization around the minima~\cite{pittorino2020entropic, jiang2019fantastic}. Each point in this plot shows the mean over five networks trained from different initializations. When sparsification is applied while increasing network depth, the maximum accuracy and robustness are achieved for smaller depth values 
in all experiments. Note that keeping a network connected while increasing its depth, in contrast to width, becomes difficult with higher sparsity. The results are summarized in \Secref{sec:depth_experiments}. The outcome across all experiments consistently suggests that sparsification alone does not undermine network robustness to weight perturbations as long as sufficient network connectivity is maintained. 


\fakeparagraph{Corrupted data}
\Figref{fig:width:corrupted_data} evaluates the performance of the models on corrupted datasets MNIST-C, CIFAR10-C and CIFAR100-C. We observe that as we move towards higher sparsity levels, the test performance first increases then decreases in extreme sparsity levels. We note that such increase is happening earlier for simpler tasks like MNIST.
Each point in \Figref{fig:width:corrupted_data}is mean performance over three trained networks. For each network we randomly sample 1000 examples from a dataset and add five noise samples in each run. On CIFAR10-C and CIFAR100-C our evaluation considers corruption severity of two and four as classified by \cite{hendrycks2019robustness}. Detailed results for specific corruption types can be found in \Secref{sec:corrupted_data_details}. The results for the achieved performance of networks sparsified by increasing depth are also shown in \Secref{sec:depth_experiments}. 
We note that VGG networks experience convergence issues as the network sparsity approaches 10\% due to lacking connectivity between layers. This is not the case for MLP and ResNet which also converge for lower percentage of remaining weights. We attribute these differences to the power of skip connections in ResNet and low overall tested network depths (1,2,4 and 8) for MLP.

\fakeparagraph{Sensitive classes}
Similarly to \cite{hooker2020compressed}, sensitive classes are considered those with the lowest recall. For each sparsity level, we train five models, evaluate them on the test data and report the minimum recall among all classes. \cite{hooker2020compressed} shows that there are some particular examples in each class that a pruned network forgets easily. However, we observe that as the networks get wider (or deeper) and sparser the minimum recall does not decrease. Sparsification does not disproportionally affect sensitive classes, which may not be noticeable by just looking at the overall accuracy. This is due to the fact that the capacity of the networks is fixed.  The results are shown in \Figref{fig:perturbated_models_recall} in \Secref{sec:depth_experiments}.

\fakeparagraph{Adversarial attacks}
\Figref{fig:adv_corrupted_data_selected} shows the robustness of sparsified networks when applying adversarial attacks to perturb test data. We observe a consistent trend for robustness to all adversarial attacks (BIM~\cite{kurakin2016adversarial}, APGD~\cite{croce2020reliable}, PGD~\cite{madry2019deep}, FGSM~\cite{goodfellow2014explaining}). Similar to perturbed model weights and corrupted data, as we have less remaining weights, test performance for adversarial examples is first improved and then decreases for extreme sparsity levels where the overall (clean) network accuracy drops. Dense VGG networks trained on MNIST show the highest accuracy decline in the presence of all attacks, while sparsification helps to improve adversarial robustness.

\fakeparagraph{Post-training sparsification}
\Figref{fig:posttraining:perturbated_models} depicts the results for post-training sparsification for MLP and VGG architectures challenged with perturbed model weights. The results indicate a similar trend to the experiments with static sparsity applied at initialization. For VGG we observe a slight improvement followed by an accuracy drop. However, the performance does deteriorate sooner than with static sparsity. For MLP the results show stagnating accuracy and a slight drop in performance on CIFAR-10. We attribute this to the simplicity of our sparsification method and a relatively low number of weights in the one layer MLP. Similar results are obtained on corrupted datasets visualized in \Figref{fig:posttraining:corrupted_data}.

\section{Conclusion}
\label{sec:conclusion}

In this work we hypothesise that sparsity, while keeping the number of parameters fixed, does not hurt network robustness. We provide experimental evidence to support this claim based on several standard architectures, datasets, sparsification methods and measures of robustness. Our observation is that network sparsification often helps to improve robustness compared to a dense model, yet the benefits decline together with the overall model accuracy for high sparsity levels. This is due to the increasingly loose connectivity between layers which complicates optimization. Since network capacity rather than sparsity causes accuracy and robustness drop of compressed models, designing pruning methods that treat network capacity and sparsity separately can lead to better compressed models. In addition, our work emphasizes the need for training procedures that better support sparse operations, which would allow for a faster and more memory efficient training of sparse networks.

\clearpage
\bibliography{references}
\bibliographystyle{icml2021}

\appendix
\section{Implementation Details} \label{sec:implementationdetails}
We used Caliban~\cite{ritchie2020caliban} to manage all experiments in a reproducible environment in Google Cloud’s AI Platform. 
Each point in plots show the mean value taken over at least five different runs.

\subsection{Training for MLP} \label{sec:mlp-training}
\begin{itemize}
    \item Dataset: MNIST~\cite{mnist}, CIFAR-10~\cite{cifar100} and CIFAR-100~\cite{cifar100}
    \item Network: MLP
    \item Width experiments: single hidden layer, hidden neurons $2^{n}$ where n $\in {7..15} $. 
    \item Depth experiments: $2^{9}$ neurons per layer, we add additional layers with the same number of neurons.
    \item Hyper parameters:
    \begin{itemize}
        \item LR = fixed 0.01
        \item stopping criteria = 300 epochs or loss (CE) $<$ 0.01
        \item Momentum = 0.9
    \end{itemize}
    
\end{itemize}

\subsection{Training for CNN architecture family} \label{sec:sconv-training}
\begin{itemize}
    \item Dataset: MNIST~\cite{mnist}, CIFAR-10~\cite{cifar100} and  CIFAR-100~\cite{cifar100}
    \item Network: VGG~\cite{simonyan2014very}, ResNet~\cite{he2015deep})
    \item Width experiments: we use VGG11 and ResNet18 changing the width of the first layer and adapting the following layers according to the ratios in the vanilla version of the networks. 
    \item Depth experiments: we change the architecture: VGG11, VGG13, VGG16 and ResNet18, ResNet34, ResNet50 while keeping the default width of the vanilla networks.

    \item Hyper parameters:
    \begin{itemize}
        \item LR = Cosine Annealing~\cite{loshchilov2016sgdr} with initial LR=0.01
        \item stopping criteria = 300 epochs or loss $<$ 0.1
        \item Momentum = 0.9
    \end{itemize}
    
\end{itemize}

\section{Additional Plots}

\subsection{Sparsification by increasing network depth}
\label{sec:depth_experiments}

In addition to increasing network width before applying random static sparsification, we also test sparsification by increasing network depth. In this case, we duplicate layers while keeping their width unchanged. In contrast to wider networks, high sparsity levels severely impairs connectivity between layers of deeper networks which make these more difficult to train and converge. VGG experience convergence issues when sparsifying by increasing depth when sparsity approaches 10\,\%. ResNet appears less susceptible to this issue due to the presence of skip connections, whereas MLPs are more robust due to a considerably lower depth in our settings (1,2,4 and 8). Our experiments with deeper MLP networks and high sparsity also reveal convergence issues similarly to VGG. We run experiments for increasing depth for all architectures and datasets up to a point where the overall network accuracy starts to drop and as long as network training converges.

\begin{figure*}[t]
    \centering
    \begin{subfigure}[b]{0.33\linewidth}
        \centering
        \includegraphics[width=\linewidth]{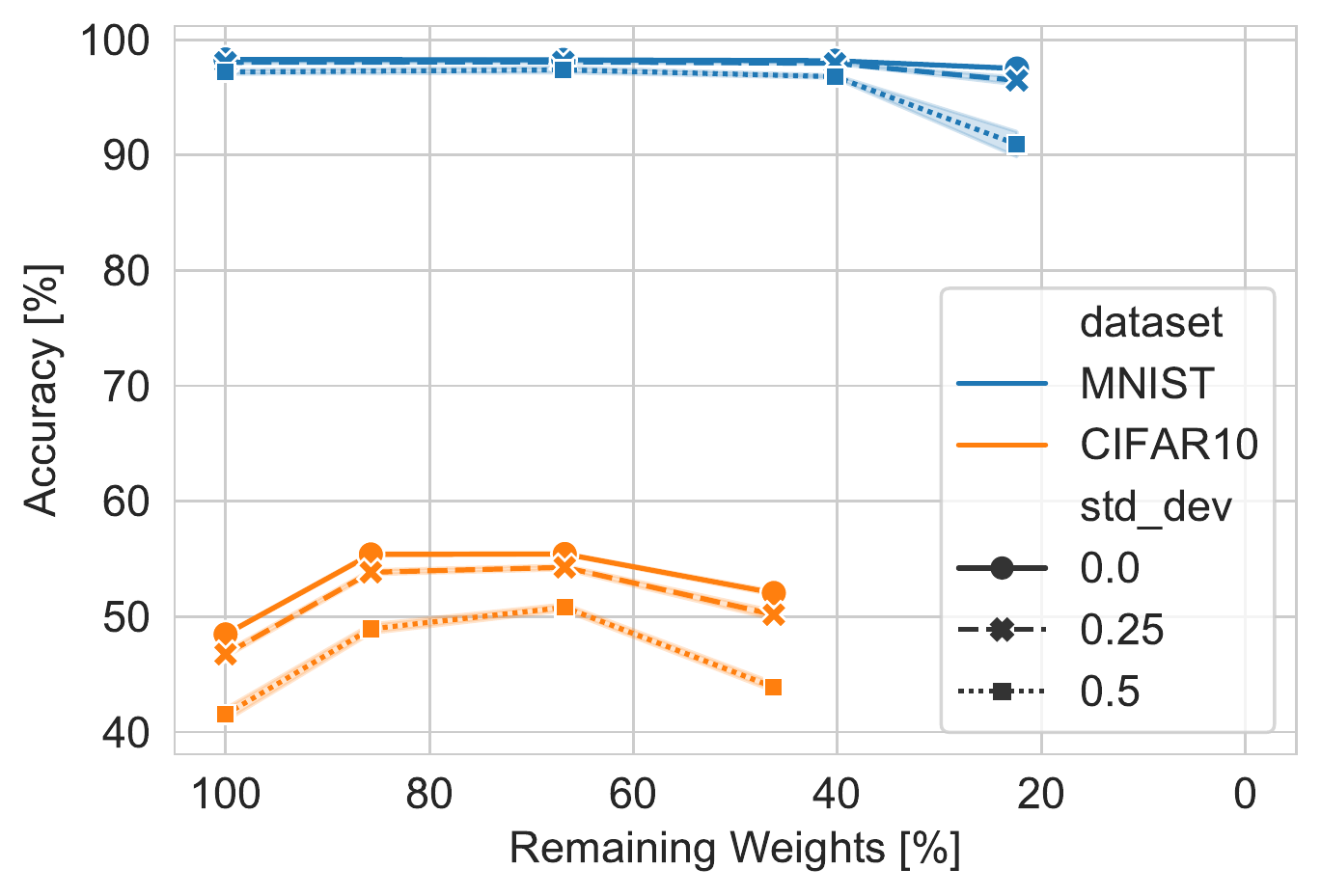}
        \caption{MLP}
        \label{fig:depth:perturbated_models_mlp}
    \end{subfigure}
    \hskip 3cm
    \begin{subfigure}[b]{0.33\linewidth}
        \centering
        \includegraphics[width=\linewidth]{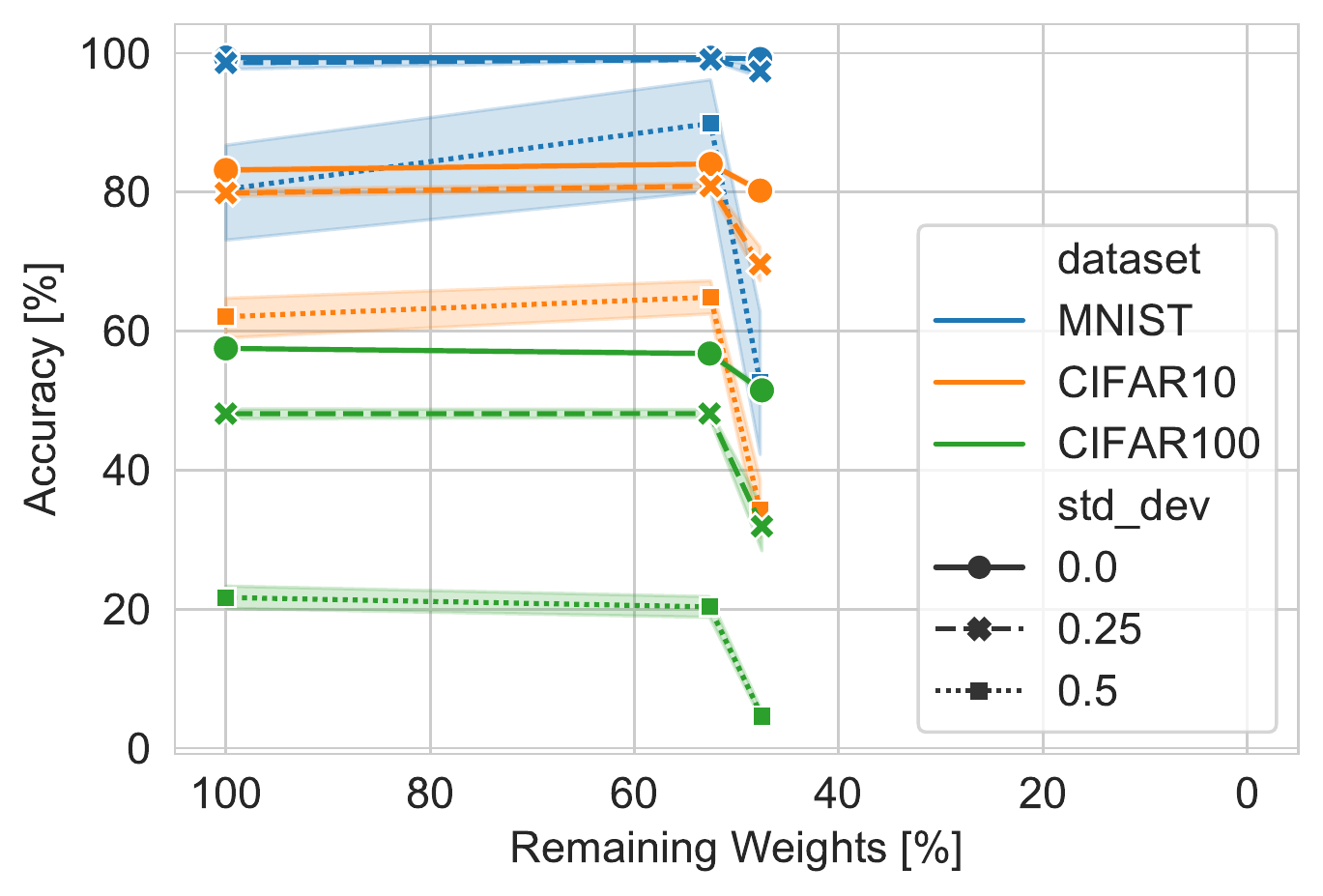}
        \caption{ResNet}
        \label{fig:depth:perturbated_models_resnet}
    \end{subfigure}
    \caption{{\bf Robustness to weight perturbations.   Sparsification by increasing network depth} We add multiplicative Gaussian noise$z_i\sim\gN(\mu, w_i^2\sigma_i^2)$ to each weight and evaluate model performance. There is a sweetspot corresponding to optimal sparsity. With higher depth and sparsity network connectivity declines leading to simultaneous accuracy and robustness drop.}
    \label{fig:depth:perturbated_models}
\end{figure*}

\begin{figure*}[t]
    \centering
    \begin{subfigure}[b]{0.33\linewidth}
        \centering
        \includegraphics[width=\linewidth]{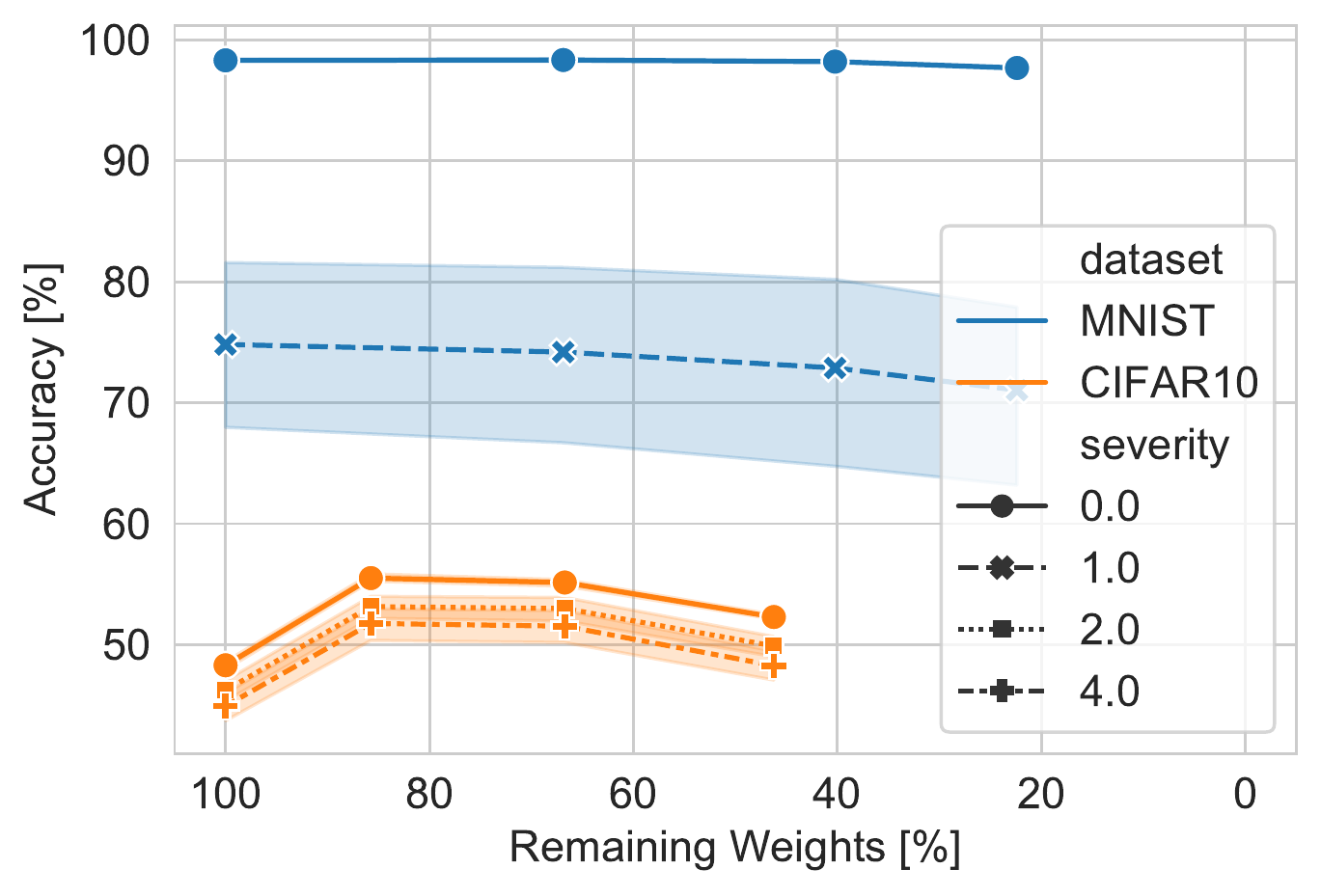}
        \caption{MLP}
        \label{fig:depth:corrupted_data_mlp}
    \end{subfigure}
    \hskip 3cm
    \begin{subfigure}[b]{0.33\linewidth}
        \centering
        \includegraphics[width=\linewidth]{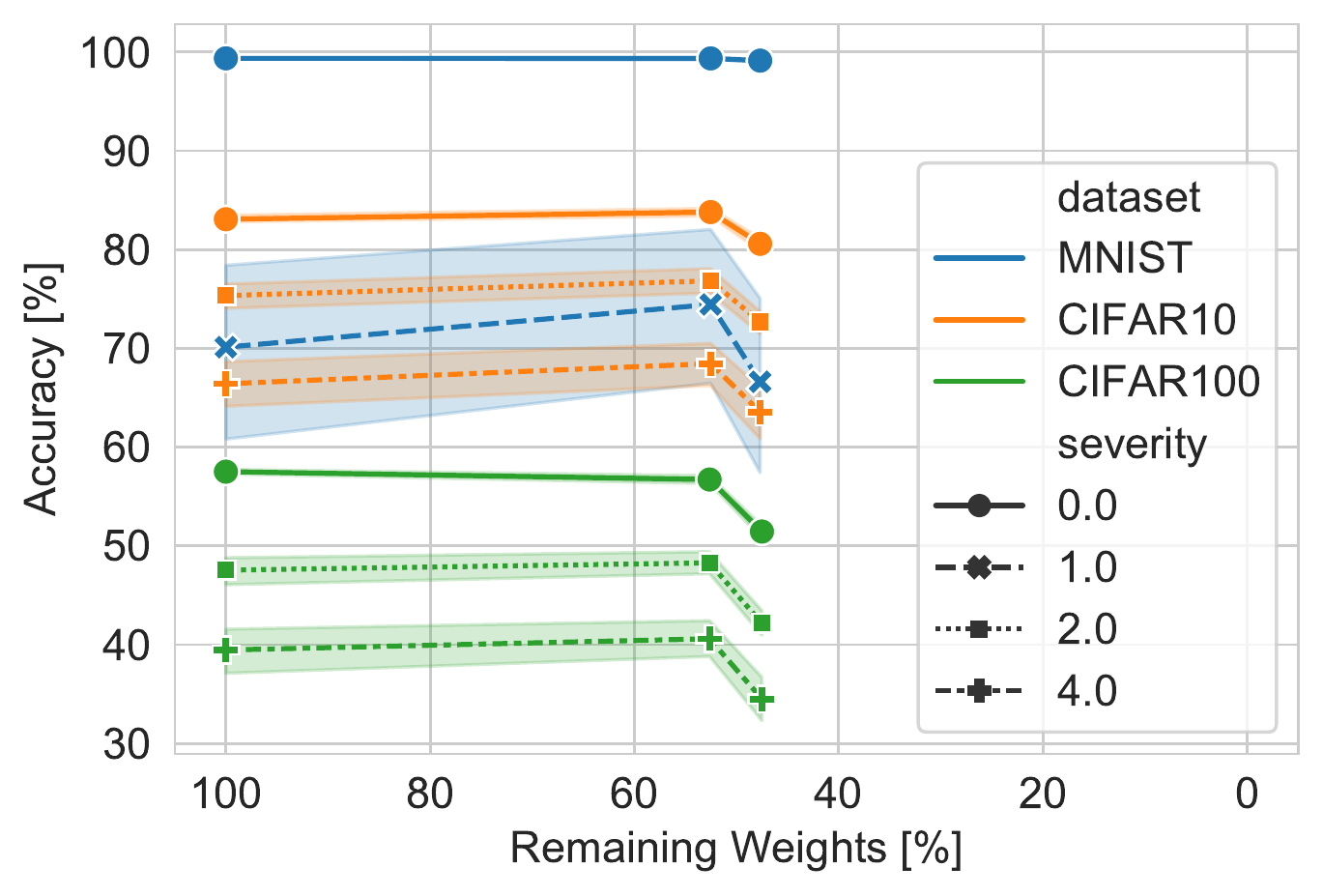}
        \caption{ResNet}
        \label{fig:depth:corrupted_data_resnet}
    \end{subfigure}
    \caption{{\bf Robustness to corrupted data.   Sparsification by increasing network depth} Corrupted datasets MNIST-C, CIFAR10-C and CIFAR100-C. There is a sweetspot corresponding to optimal sparsity. With higher depth and sparsity network connectivity declines leading to simultaneous accuracy and robustness drop.}
    \label{fig:depth:corrupted_data}
\end{figure*}

\Figref{fig:depth:perturbated_models} and \Figref{fig:depth:corrupted_data} show the results for weight perturbations and corrupted data of varying severity. The results are similar and show that robustness improves as long as network accuracy without intervention remains steady.

\begin{figure*}[h]
    \centering
    \subfloat[Sparsification by increasing network width]{
        \includegraphics[width=.33\linewidth]{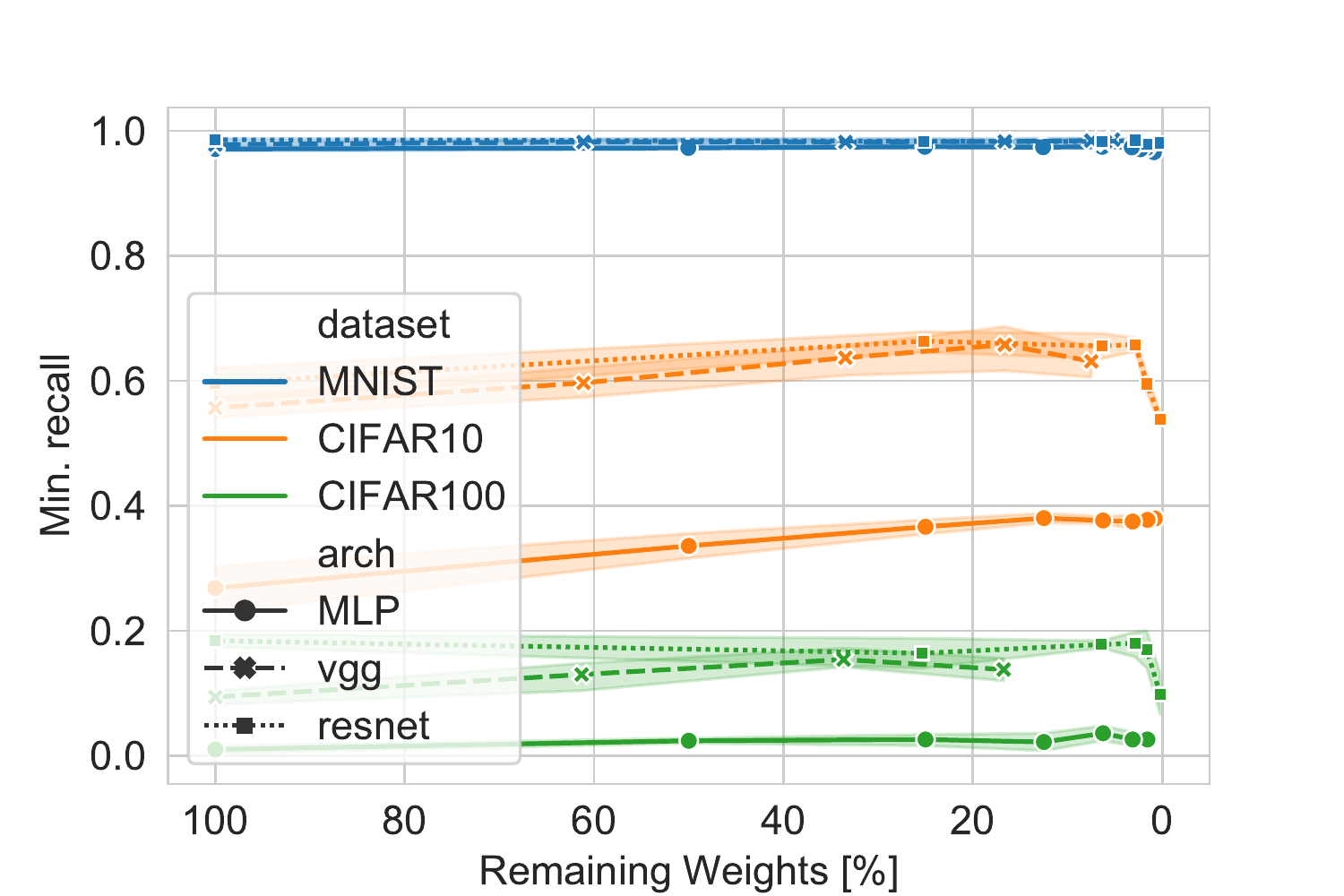}
    }
    \hskip 3cm
    \subfloat[Sparsification by increasing network depth]{
        \includegraphics[width=.33\linewidth]{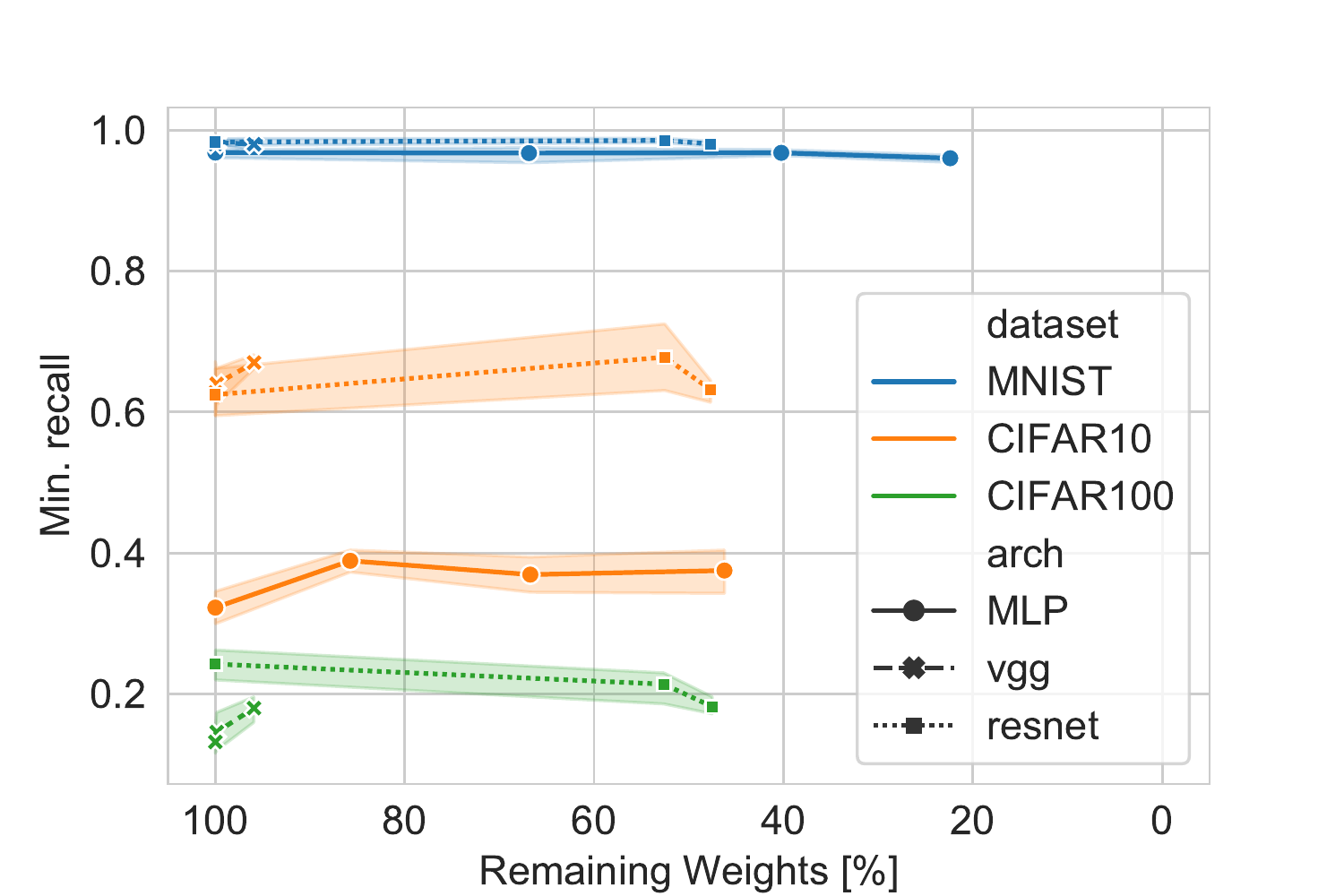}       
    }
    \caption{{\bf Minimum recall among all classes.} Network accuracy on the most sensitive classes does not decline with sparsity. Tested without weight / data corruptions. Sparsification by increasing network width (left) and depth (right).}
    \label{fig:perturbated_models_recall}
\end{figure*}

\Figref{fig:perturbated_models_recall} shows the results for the lowest recall among all classes. Network accuracy on the most sensitive classes does not decline with sparsity. The test is conducted without weight perturbation or data corruption.

\subsection{Detailed results for corrupted data}
\label{sec:corrupted_data_details}

\begin{figure*}[htb]
    \centering
    \begin{subfigure}[b]{0.33\textwidth}
        \centering
        \includegraphics[width=\linewidth]{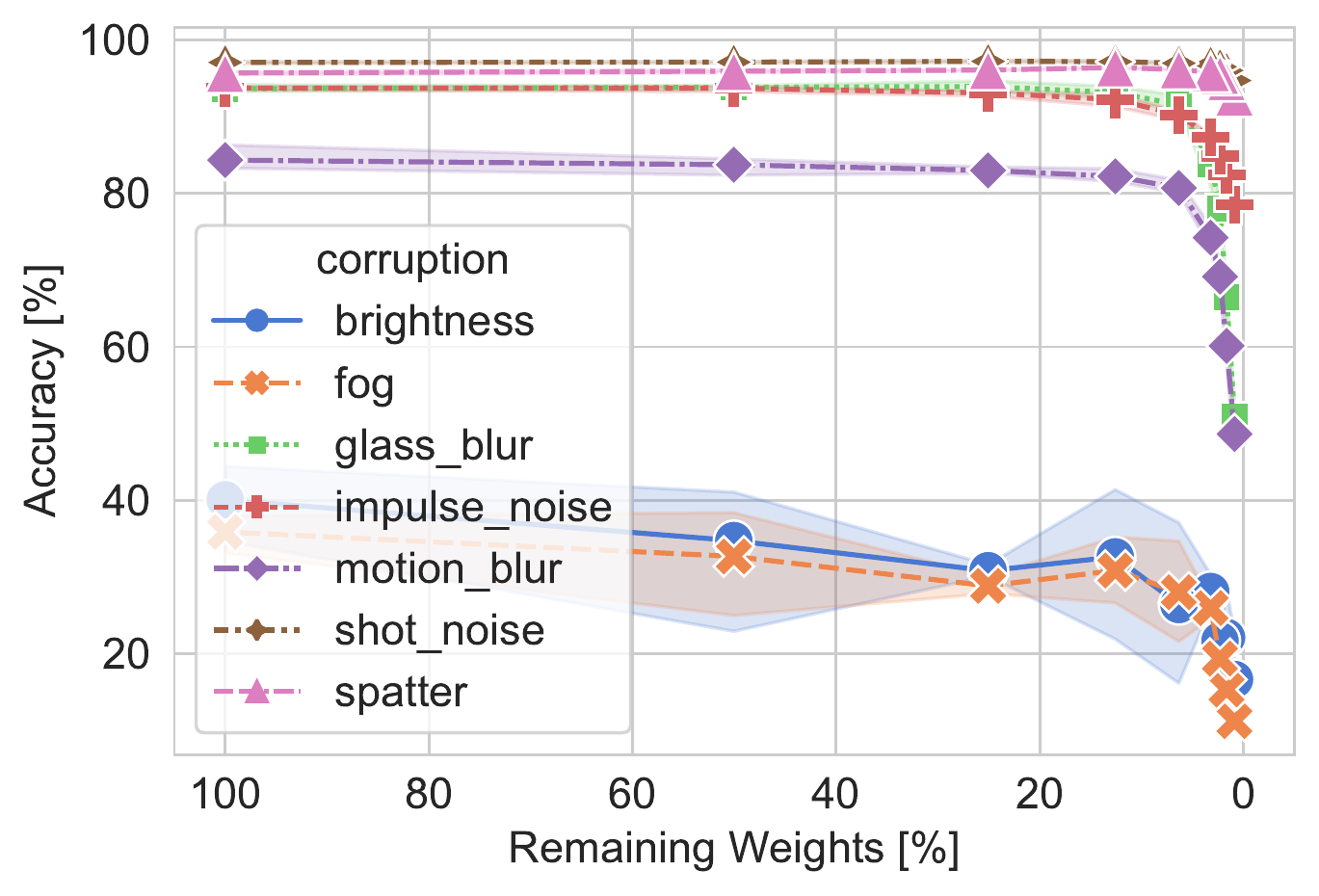}
        \caption{MNIST-C}
    \end{subfigure}
    \hfill
    \begin{subfigure}[b]{0.33\textwidth}
        \centering
        \includegraphics[width=\linewidth]{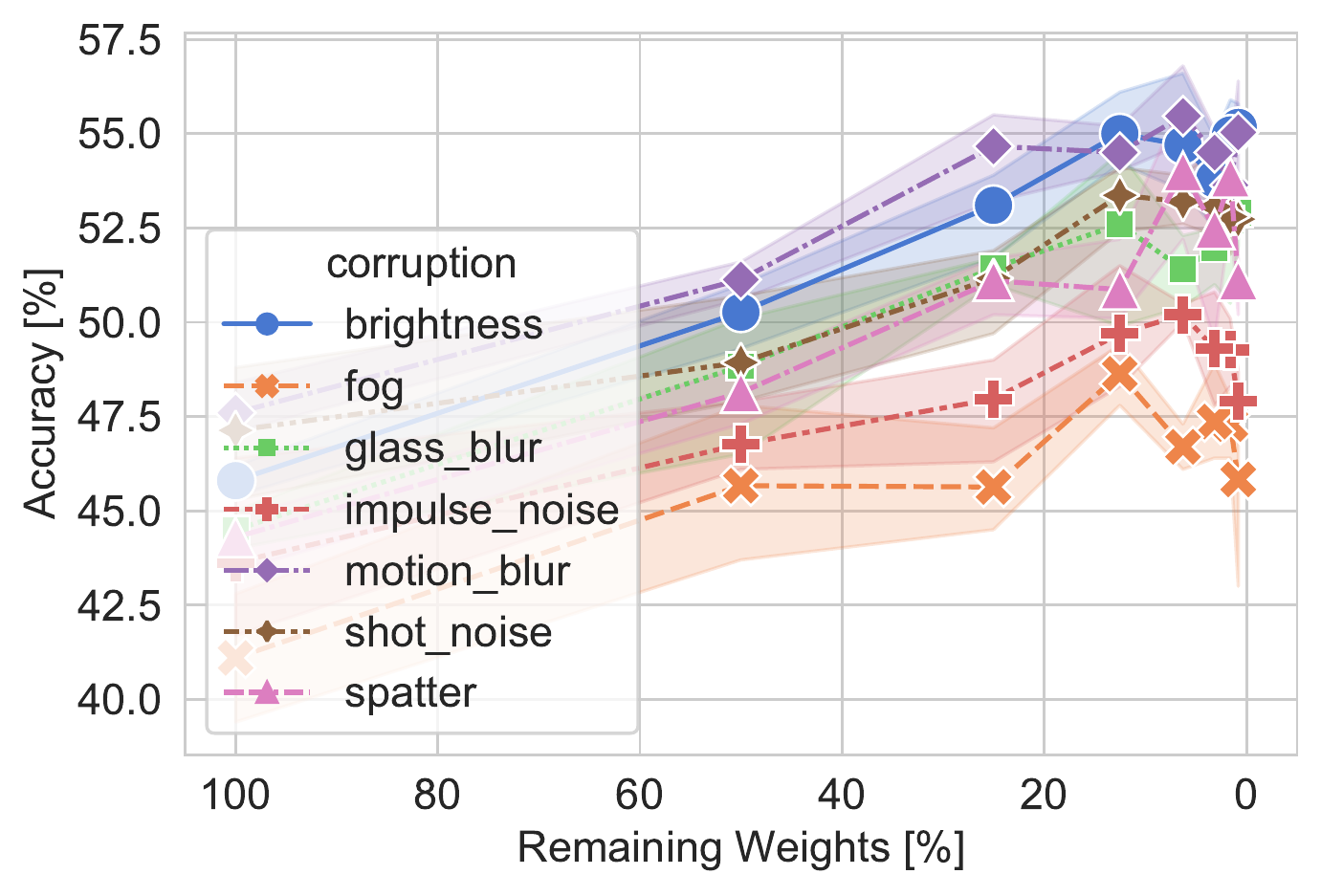}
        \caption{CIFAR10-C}
    \end{subfigure}
    \hfill
    \begin{subfigure}[b]{0.33\textwidth}
        \centering
        \includegraphics[width=\linewidth]{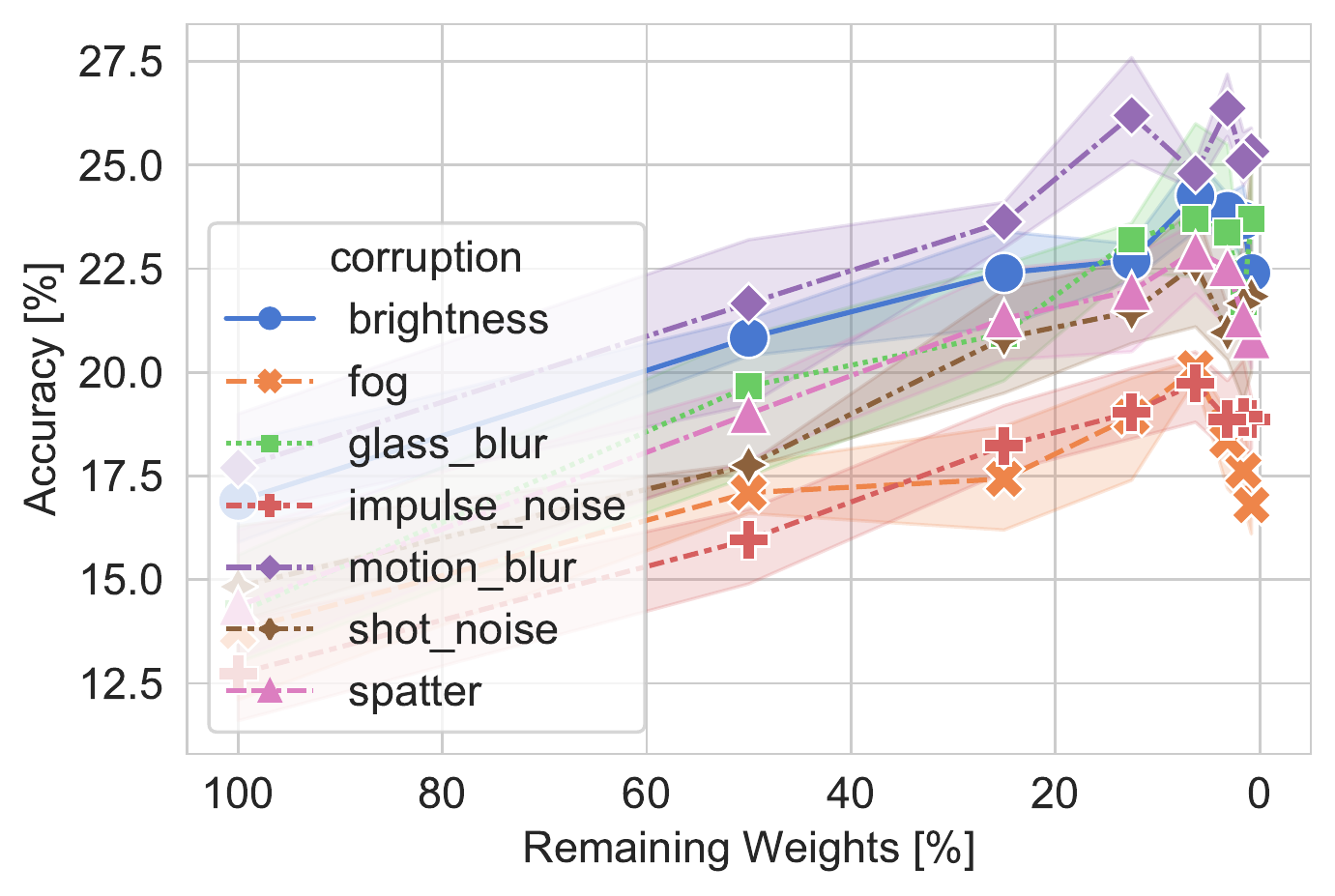}
        \caption{CIFAR100-C}
    \end{subfigure}
    \caption{{\bf One layer MLP performance on selected corruption types.} For CIFAR10-C and CIFAR100-C we observe a clear trend across all corruption types, which suggests that the sparser networks with increased width are more robust. We note that for simpler task MNIST-C such increase in the performance is happening earlier in sparsity levels.}
    \label{fig:corrupted_data_mlp_selected}
\end{figure*}

\begin{figure*}[t]
    \centering
    \begin{subfigure}[b]{0.33\textwidth}
        \centering
        \includegraphics[width=\linewidth]{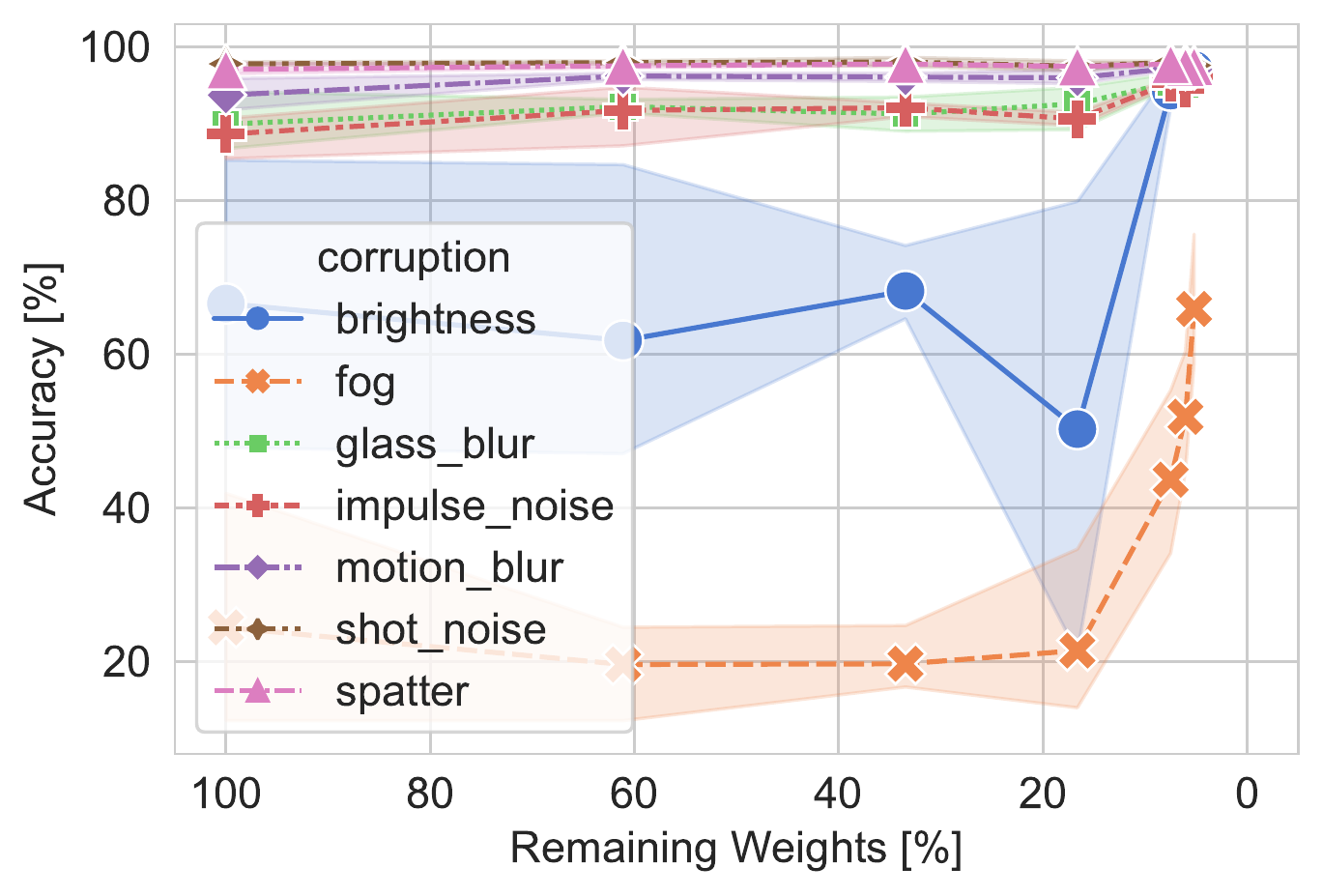}
        \caption{MNIST-C}
    \end{subfigure}
    \hfill
    \begin{subfigure}[b]{0.33\textwidth}
        \centering
        \includegraphics[width=\linewidth]{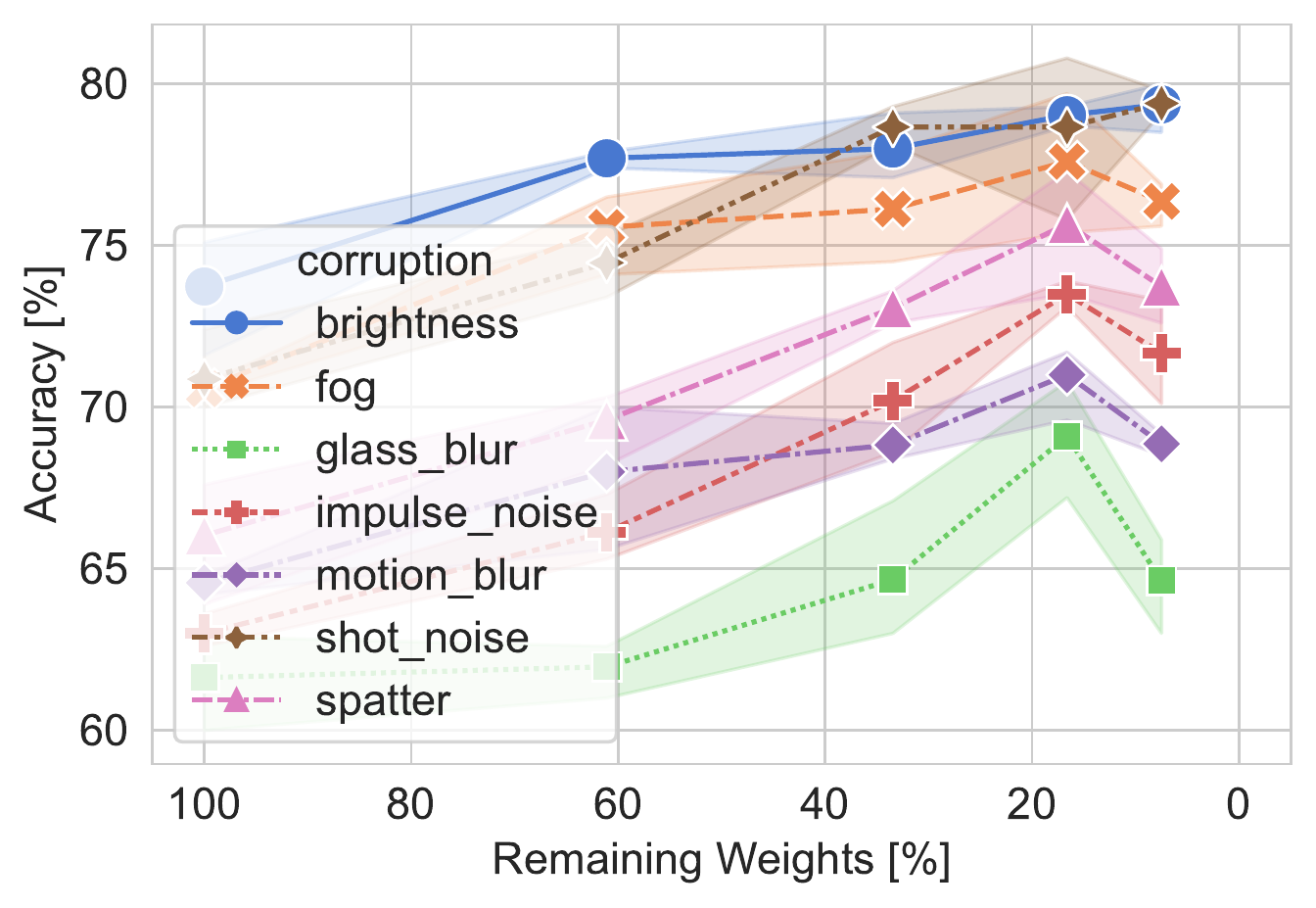}
        \caption{CIFAR10-C}
    \end{subfigure}
    \hfill
    \begin{subfigure}[b]{0.33\textwidth}
        \centering
        \includegraphics[width=\linewidth]{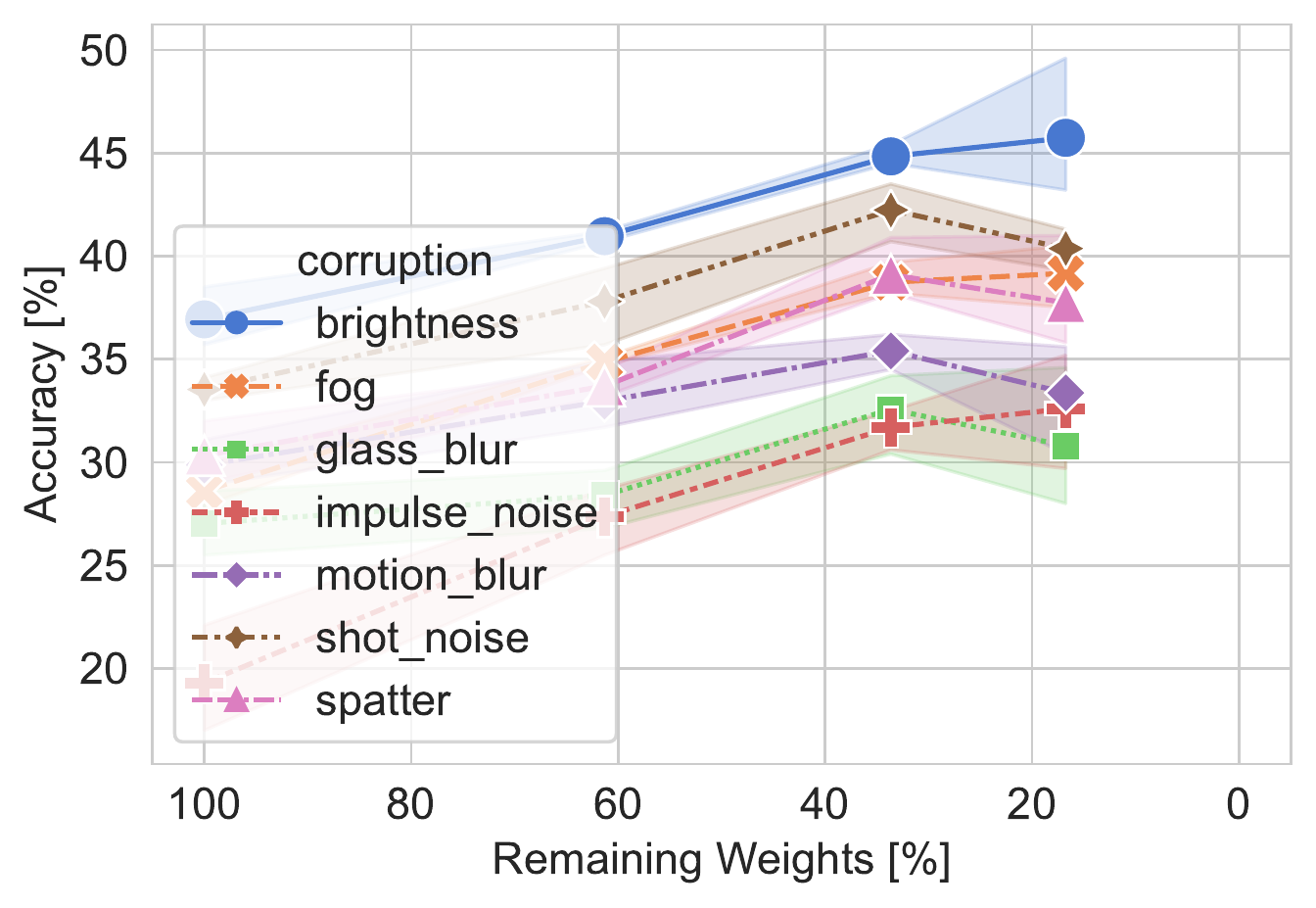}
        \caption{CIFAR100-C}
    \end{subfigure}
    \caption{{\bf VGG11 performance on selected corruption types.} The results show a clear upwards trend across different corruption types which indicates, that the networks get more robust as the sparsity and width increase.}
    \label{fig:corrupted_data_vgg_selected}
\end{figure*}

\begin{figure*}[t]    
    \begin{subfigure}[b]{0.33\textwidth}
        \centering
        \includegraphics[width=\linewidth]{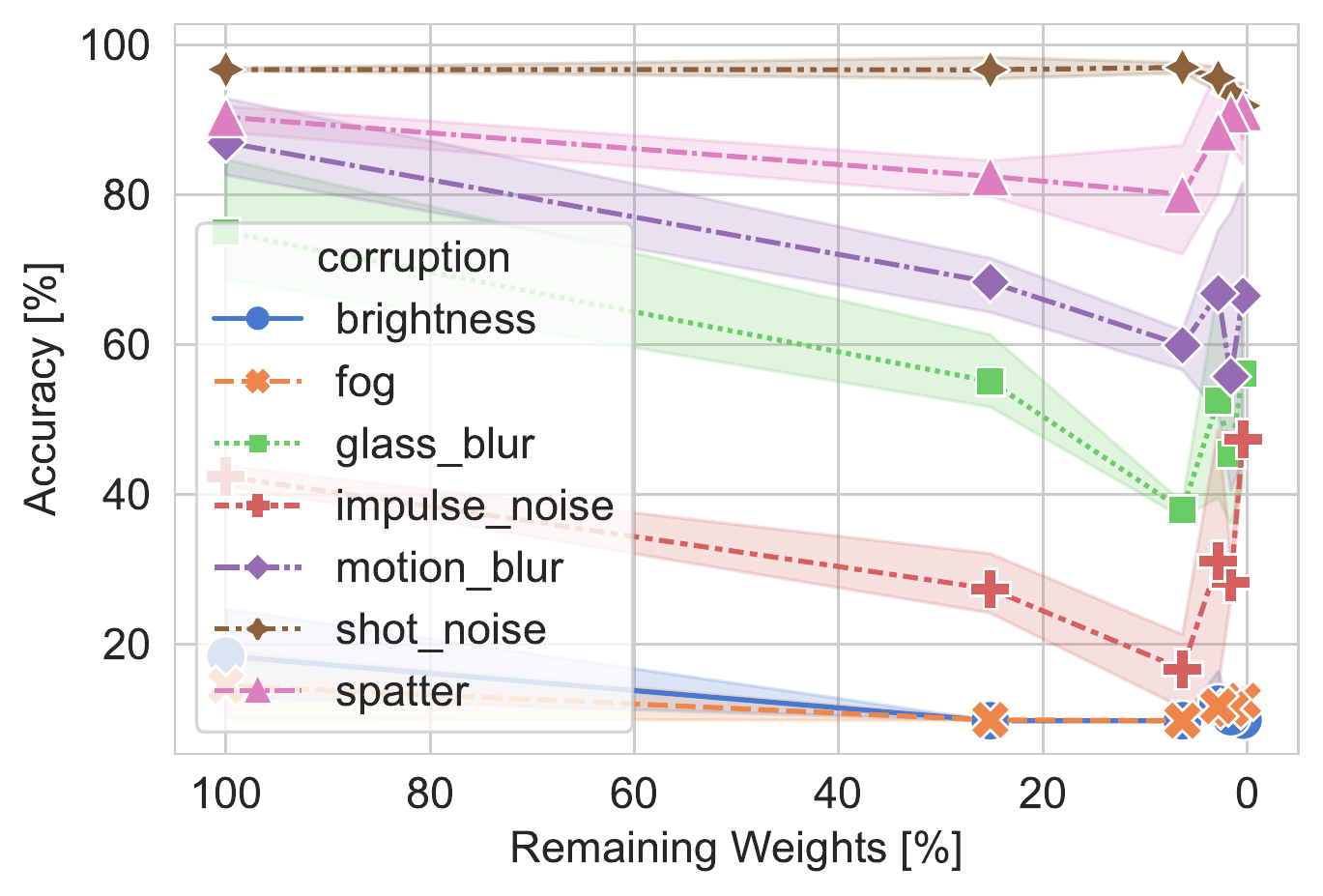}
        \caption{ MNIST-C}
    \end{subfigure}
    \hfill
    \begin{subfigure}[b]{0.33\textwidth}
        \centering
        \includegraphics[width=\linewidth]{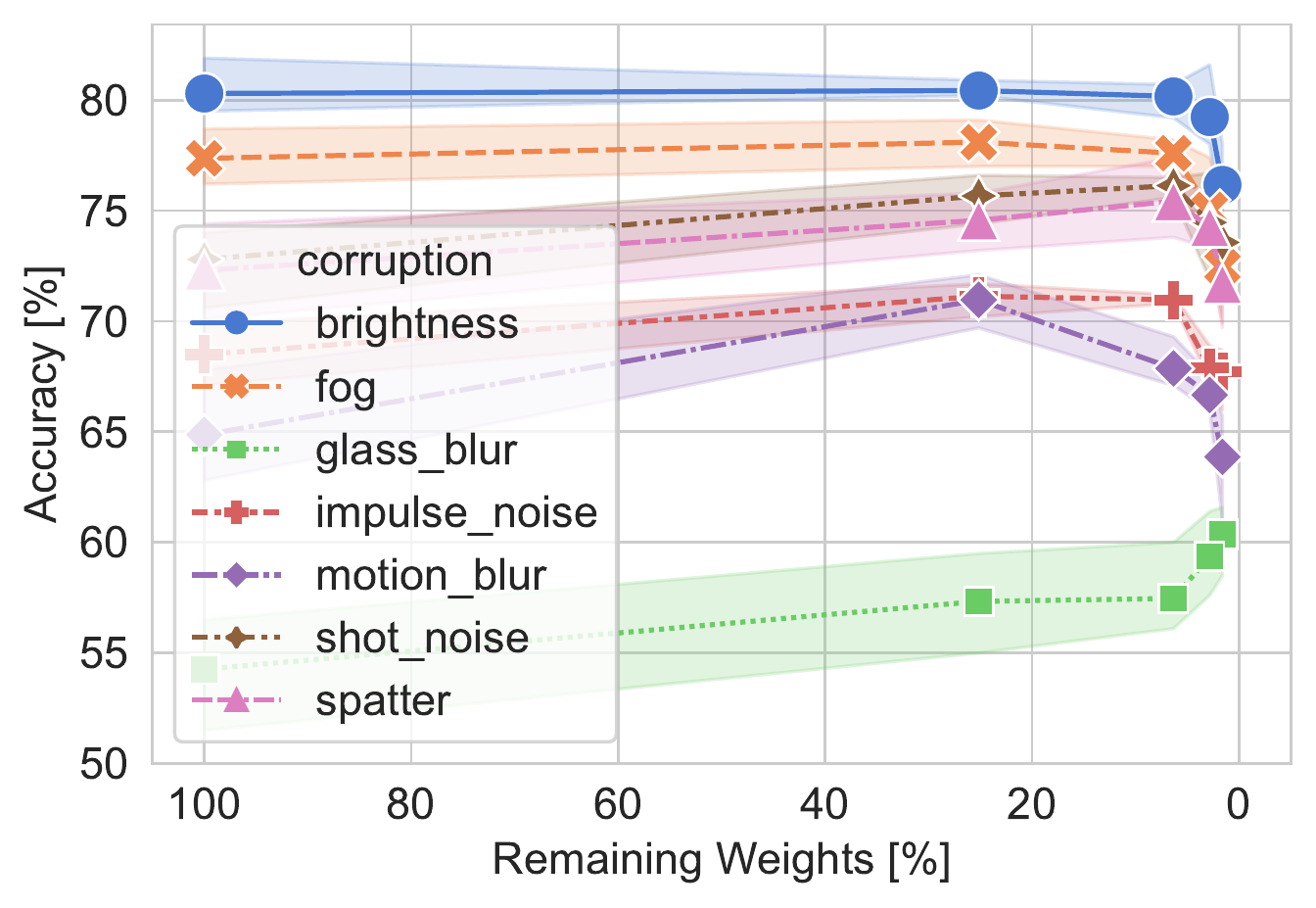}
        \caption{ CIFAR10-C}
    \end{subfigure}
    \hfill
    \begin{subfigure}[b]{0.33\textwidth}
        \centering
        \includegraphics[width=\linewidth]{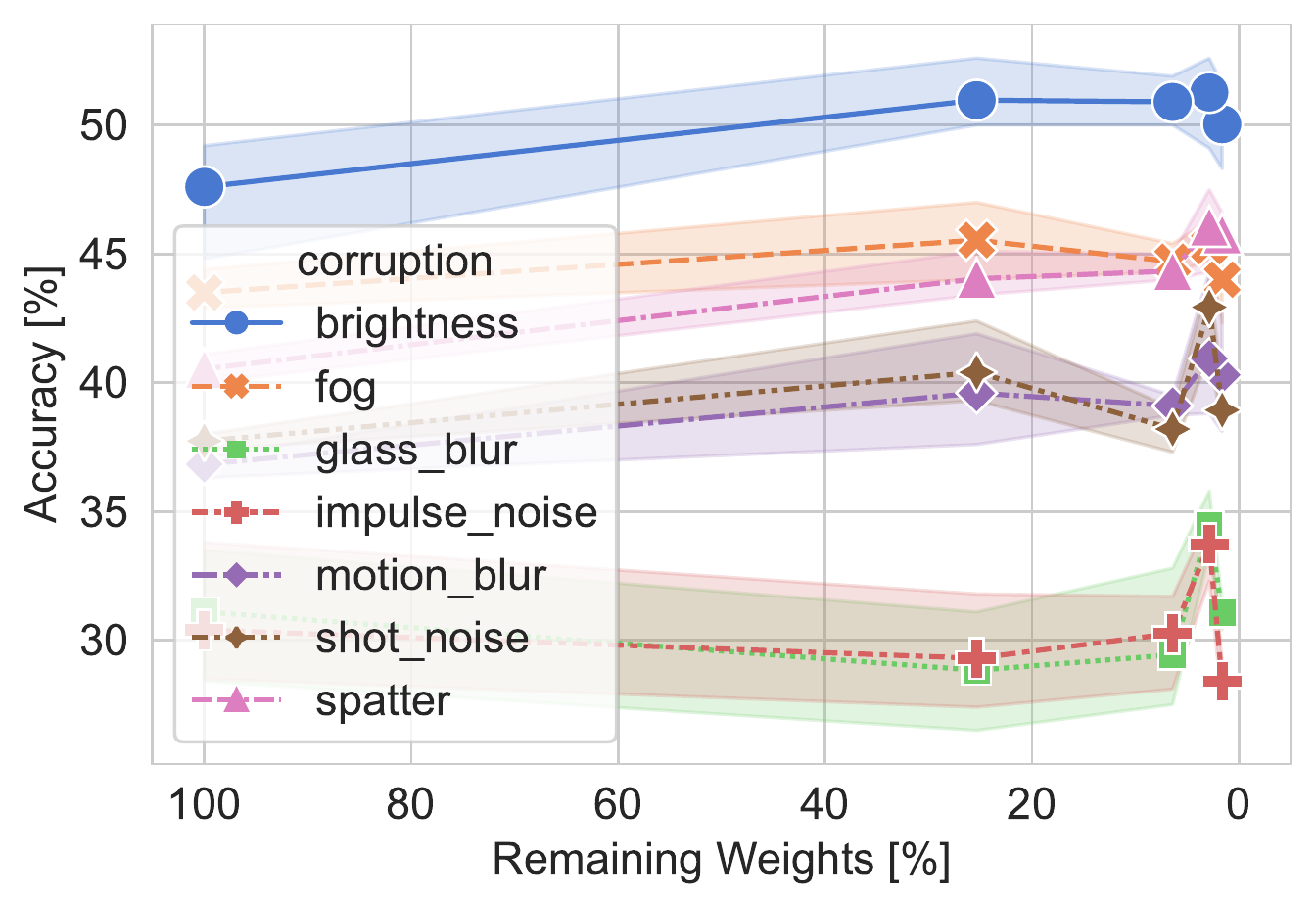}
        \caption{CIFAR100-C}
    \end{subfigure}
    \caption{{\bf ResNet18 performance on selected corruption types.} We observe a upwards trend across corruption types for CIFAR10-C and CIFAR100-C, models with higher width and higher sparsity perform better on corrupted data. We note that the increase in the performance for simpler task MNIST-C happens sooner.}
    \label{fig:corrupted_data_resnet_selected}
\end{figure*}

The plots presented in \Figref{fig:corrupted_data_mlp_selected}, \Figref{fig:corrupted_data_vgg_selected} and \Figref{fig:corrupted_data_resnet_selected} provide details on the impact of individual corruption methods on network performance. We show details for MLP, VGG and ResNet architectures trained on MNIST, CIFAR-10 and CIFAR-100. Sparsification is achieved by increasing network width. Although the effect of data corruptions on model performance varies widely, it can be observed that in all cases a sparser network matches the accuracy of the vanilla 100\,\% network. This observation holds up to high sparsity levels where the overall model performance declines.

\end{document}